\documentclass[sigconf]{acmart}
\pdfoutput=1
\AtBeginDocument{%
  \providecommand\BibTeX{{%
    \normalfont B\kern-0.5em{\scshape i\kern-0.25em b}\kern-0.8em\TeX}}}

\copyrightyear{2021}
\acmYear{2021}
\setcopyright{acmcopyright}\acmConference[KDD '21]{Proceedings of the 27th ACM SIGKDD Conference on Knowledge Discovery and Data Mining}{August 14--18, 2021}{Virtual Event, Singapore}
\acmBooktitle{Proceedings of the 27th ACM SIGKDD Conference on Knowledge Discovery and Data Mining (KDD '21), August 14--18, 2021, Virtual Event, Singapore}
\acmPrice{15.00}
\acmDOI{10.1145/3447548.3467379}
\acmISBN{978-1-4503-8332-5/21/08}

\usepackage{multirow}
\usepackage[export]{adjustbox}
\usepackage[outdir=./]{epstopdf}

\usepackage{graphicx}
\usepackage{enumerate}
\usepackage{textcomp}
\usepackage{xcolor}
\usepackage{url}
\usepackage{balance}
\usepackage{mathrsfs}

\usepackage{graphicx}
\usepackage{textcomp}
\usepackage{xcolor}

\theoremstyle{definition}
\newtheorem{definition}{Definition}[section]

\usepackage{algorithm} 
\usepackage{algpseudocode}
\newcommand{\etal}{\textit{et al.}}

\usepackage[tight,footnotesize]{subfigure}

\newcommand{\eat}[1]{}
\newcommand{\vsa}{\vspace*{-0.28cm}}
\newcommand{\vsb}{\vspace*{-0.19cm}}
\newcommand{\vsc}{\vspace*{-0.16cm}}

\usepackage{color,soul,xspace}
\newcommand{\todo}[1]{\sethlcolor{red}\hl{TODO: #1}} 

\DeclareMathAlphabet\mathbfcal{OMS}{cmsy}{b}{n}

\newcommand{\mourmeth}{\text{TGSD}}
\newcommand{\ourmeth}{$\mourmeth$\xspace}

\begin{document}

 \fancyhead{}

\title{Temporal Graph Signal Decomposition}

\author{Maxwell McNeil}
\authornote{Corresponding Author}
\email{mmcneil2@albany.edu}
\affiliation{%
  \country{University at Albany---SUNY, USA}
}

\author{Lin Zhang}
\email{lzhang22@albany.edu}
\affiliation{%
  \country{University at Albany---SUNY, USA}
}
\author{Petko Bogdanov}
\email{pbogdanov@albany.edu}
\affiliation{%
  \country{University at Albany---SUNY, USA}
}


\begin{abstract}
Temporal graph signals are multivariate time series with individual components associated with nodes of a fixed graph structure. Data of this kind arises in many domains including activity of social network users, sensor network readings over time, and time course gene expression within the interaction network of a model organism. Traditional matrix decomposition methods applied to such data fall short of exploiting structural regularities encoded in the underlying graph and also in the temporal patterns of the signal. How can we take into account such structure to obtain a succinct and interpretable representation of temporal graph signals?  

We propose a general, dictionary-based framework for temporal graph signal decomposition (TGSD). The key idea is to learn a low-rank, joint encoding of the data via a combination of graph and time dictionaries. 
We propose a highly scalable decomposition algorithm for both complete and incomplete data, and demonstrate its advantage for matrix decomposition, imputation of missing values, temporal interpolation, clustering, period estimation, and rank estimation in synthetic and real-world data ranging from traffic patterns to social media activity. 
Our framework achieves $28\%$ reduction in RMSE compared to baselines for temporal interpolation when as many as $75\%$ of the observations are missing. It scales best among baselines taking under 20 seconds on 3.5 million data points and produces the most parsimonious models. To the best of our knowledge, TGSD is the first framework to jointly model graph signals by temporal and graph dictionaries.
\end{abstract}



\maketitle


\section{Introduction}
Multivariate time series often feature temporal and ``spatial'' structure inherent to the domain in which they are collected. Incorporating this structure when mining such data is essential to obtaining parsimonious, robust, and interpretable models. In this paper we focus specifically on temporal data associated with the nodes of a fixed graph and refer to such data as \emph{temporal graph signals (TGS)}. Example settings abound: traffic over time in transportation networks~\cite{joshi2015review}, temporal readings in sensor networks~\cite{Modeling15-Chen}, gene expression time series overlaid on protein interactions networks~\cite{bar2004analyzing}, and user opinion dynamics in social networks~\cite{amelkin2017distance}. Beyond structural coupling on the graph, temporal graph signals may exhibit temporal structures such as periodicity, smoothness, trends, and others.
Our goal in this work is to obtain a low-dimensional embedding of a TGS which jointly exploits such graph and temporal properties. It is important to note that the TGS setting is fundamentally different from dynamic graph mining where the structure (sets of nodes and edges) evolves as opposed to signals over the nodes~\cite{borgwardt2006pattern}. 
\begin{figure}[t]
    \footnotesize
    \centering
    \includegraphics[trim={0cm 9cm 3cm 1cm},clip,width=0.47\textwidth]{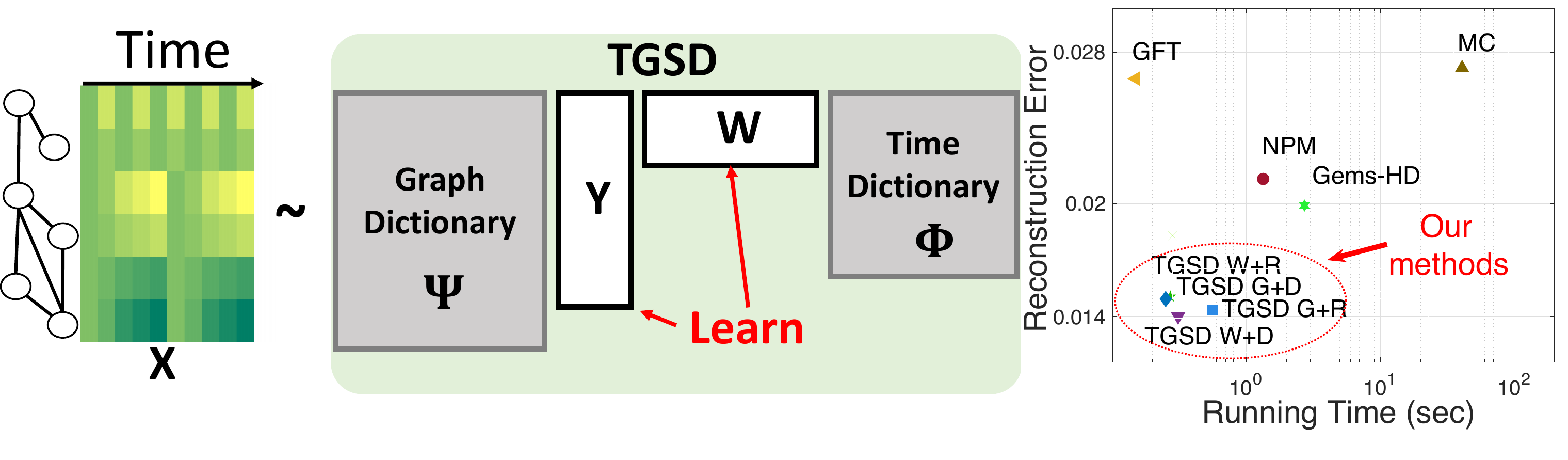} \vsa\vsb
    \caption{ \footnotesize
    (Left panel:) TGSD decomposes a temporal graph signal as a product of two fixed dictionaries and two corresponding sparse encoding matrices. (Right panel:) Variants of TGSD (lower left corner) are both more accurate in missing value reconstruction and faster than most baselines from the literature. Details of competing techniques are available in Sec.~\ref{sec:exp}.  }
    \label{fig:overview}\vsa\vsb
\end{figure}
The benefit of employing knowledge of the underlying graph structure has been demonstrated in graph signal processing~\cite{ortega2018graph,Shuman_2013,dong2019learning}, where node values are treated as a signal over the graph and the spectrum of the graph Laplacian is employed as basis for reconstructing the signal. Temporal extensions have also been recently proposed~\cite{findingGems2019}. Graph structure modeling has found valuable applications in compression/summarization of node attributes~\cite{silva2015hierarchical}, in-network regression and optimization~\cite{hallac2015network}, missing value imputation~\cite{huisman2009imputation}, community detection~\cite{gorovits2018larc} and anomaly detection~\cite{akoglu2015graph}.

Similarly, temporal structures such as periodicity~\cite{tennetiTSP2015}, trends \cite{zhang2005neural} and smoothness~\cite{chakrabarti2006evolutionary} have a long history in research on time series. 
Most existing methodology focuses on modeling either the graph structure~\cite{silva2015hierarchical,hallac2015network,huisman2009imputation,akoglu2015graph} or the structure of the temporal signal~\cite{tennetiTSP2015,zhang2005neural}. The interplay between temporal and structural graph properties gives rise to important behaviors, however, no frameworks currently exist to facilitate a joint representation. For example, traffic levels in a transportation network are shaped by both network locality and the time of the day~\cite{dong2009network}.  

Missing values~\cite{WeiNIPS2018} and irregular sampling in the time and/or the graph domain~\cite{lorenzo2018sampling} also pose important challenges in TGS data analysis. Missing values may arise due to sensor malfunction or other kinds of data corruption; sub-sampling in the graph domain may arise in social media data due to privacy constraints, while temporal sub-sampling may arise due to limitations of how fast time snapshots can be collected. A joint structural modeling of the graph and temporal patterns can be especially advantageous in such settings of missing or incomplete data. Hence, the key question we seek to address in this paper is: \emph{How to efficiently learn a succinct, robust and interpretable representation for temporal graph signals, which can (i) jointly exploit temporal and graph structural regularities and (ii) handle missing values?}

We propose a general framework for \emph{temporal graph signal decomposition (TGSD)} based on joint dictionary encoding in the graph and time domain. TGSD decomposes the signal as a product of four matrices: two fixed dictionaries and two corresponding sparse encoding coefficient matrices (Fig.~\ref{fig:overview}, left pane). Our framework flexibly incorporates widely adopted temporal and graph dictionaries from the literature and can also be employed for data with missing values. We develop a general and efficient solver for TGSD and demonstrate its utility in datasets from different domains. The framework achieves significant performance gains over baselines for matrix decomposition, missing value reconstruction, temporal interpolation, node clustering, and periodicity detection, while scaling significantly better than most of them (Fig.~\ref{fig:overview}, right pane).   

Our contributions in this paper are as follows:

\noindent{\bf $\bullet$ Generality and novelty:} We propose a general dictionary-based decomposition framework for temporal graph signals. To the best of our knowledge \ourmeth is the first method to unify graph signal processing and time series analysis by incorporating existing as well as custom dictionaries from both domains.

\noindent{\bf $\bullet$ Scalability and parsimony:} 
TGSD scales to large instances on par or better than high-accuracy baselines. It produces low complexity and interpretable representations of the input data.

\noindent{\bf $\bullet$ Applicability and accuracy:} 
We demonstrate \ourmeth's utility for data decomposition, missing value imputation, interpolation, clustering, period detection, and rank estimation. Its quality dominates that of baselines across applications.


\vsb
\section{Related work}

{\noindent \bf Sparse dictionary coding} aims to represent data via a sparse combination of dictionary basis. It is widely used in signal processing ~\cite{zhang2015survey}, image analysis~\cite{elad2006image}, and computer vision~\cite{wright2008robust}. Dictionaries are designed to capture the underlying structure in the data, e.g. DFT~\cite{dft_2010} and DWT~\cite{findingGems2019}. Some methods learn the dictionary from data~\cite{zheng2016efficient}.  
Our work focuses specifically on signals over graphs and can accommodate arbitrary time and graph dictionaries designed to match expected structures in the data, making the literature on dictionary encoding complementary to our approach. 

\noindent{\bf Graph signal processing (GSP)} specifically models signals over graph nodes and is a popular emerging research area in signal processing~\cite{Shuman_2013}. 
A central premise is that a graph signal can be represented as a linear combination of graph dictionary basis. The eigenvectors of the graph Laplacian are often adopted as basis in this domain~\cite{dong2019learning}.
Proposals to learn the basis from data also exist in the literature~\cite{silva2018spectral}. 
Beyond a univariate signal over nodes, a recent approach considers also multivariate graph signals~\cite{findingGems2019}. Our work focuses on evolving graph signals, and thus, can be viewed as a generalization of static GSP methods. 

A recent clustering approach, called CCTN, focuses on temporal evolution of signals over graphs to clusters nodes~\cite{liu2019coupled}. CCTN computes a sparse data representation and employs the network structure as a regularizer. Different from CCTN, our work employs the graph structure for dictionary encoding and can be applied to many downstream tasks in addition to clustering. We also outperform CCTN for clustering both in terms of scalability and accuracy.


{\noindent \bf Mining and optimization for network samples} is another relevant area which differs from classical machine learning in that features are associated with network nodes~\cite{Zhang2019dsl,Ranu:2013:MDS:2487575.2487692}.
Common to this setting is that network samples share a common structure but are modeled as independent, while in our setting they are ordered in time and this temporal order is crucial. 
An optimization framework for graph node values was recently proposed with a key premise of local smoothness~\cite{hallac2015network}. This work, however, does not consider temporal evolution of the node values and uses the graph structure for regularization rather than encoding. 

\noindent{\bf Matrix completion} techniques estimate missing values in incomplete matrices~\cite{nguyen2019low}. Some employ nuclear norm minimization~\cite{Low_Rank} 
and others low-rank semi-definite programming~\cite{mitra2010large}. These methods are developed for general matrices and are agnostic to existing column/row side information (e.g., time and graph structures) which we consider in TGSD. 
Some recent methods incorporate side information as regularizers to improve the completion quality~\cite{kalofolias2014matrix,timeVaryingGraph} or employ known temporal patterns such as periodicity~\cite{tensor_traffic_predict}.
Different from the above, we perform joint row and column structured encoding, which, as we demonstrate experimentally, outperforms general matrix completion methods. 

\noindent\textbf{{Dynamic graph mining}} methods seek to discover patterns from the evolving graph structure as opposed to node signals that evolve over a fixed structure~\cite{borgwardt2006pattern,zhang2019PERCeIDs}. Classic problems in this domain area include link prediction and recommendation~\cite{dynamic_cf} community detection~\cite{zhang2019PERCeIDs} and frequent pattern mining~\cite{borgwardt2006pattern}. All these methods focus on structural dynamics, while our work focuses on node signal evolution within a fixed structure. 

\noindent\textbf{Graph neural networks} are popular deep architectures shown to achieve state-of-the-art performance in text classification~\cite{GCN_text_class}, recommender systems~\cite{MCGNN} and other tasks~\cite{GCN_OG}. GNNs employ an input graph structure to learn representations of node values based on their network context. Some methods in this domain also consider time, but are primarily designed for forecasting~\cite{STGCN}, action recognition~\cite{spatial_temp_skel_rec}, or dynamic graph structure~\cite{DGCN}, as opposed to evolving signals over a graph.
In addition, this group of methods are often designed for specific tasks (e.g. recommendation and node classification), while our framework is general as it can be employed for an array of tasks and can incorporate various dictionaries. \vsa


\section{Problem Formulation}

A graph $G=(\mathcal{M},H)$ is a set of nodes $\mathcal{M}, |\mathcal{M}|=n$, and a weighted adjacency matrix $H$ whose non-zero entries encode the strength of the edges among corresponding nodes.
The combinatorial Laplacian matrix ${L}$ associated with a graph $G$ is defined as ${L}=F-H$, where $F$ is is a diagonal matrix of weighted node degrees $F_{ii}=\sum_q H_{ij}.$ 

A temporal graph signal (TGS) is a matrix $X \in\mathbb{R}^{n\times t}$ with $t$ time snapshots.
Our goal is to succinctly encode the signal via graph and time dictionaries, resulting in a decomposition of the form:
\begin{equation}~\label{obj_2}
\begin{aligned}
    X \approx \Psi Y W\Phi,\vsa\vsb
    \end{aligned}
\end{equation}
\noindent where $\Psi \in\mathbb{R}^{n\times m}$ is a fixed graph dictionary, $\Phi\in\mathbb{R}^{ s\times t}$ is a fixed temporal dictionary and $Y\in\mathbb{R}^{m\times k}$ and $W\in\mathbb{R}^{k\times s}$ are the corresponding encoding coefficient matrices we have to estimate. We choose to use two encoding matrices $Y$ and $W$, similar to many sparse coding and non-negative matrix factorization methods, in order to learn effectively both the graph and temporal structures and maintain separable interpretable embeddings for nodes and time snapshots. The internal dimension of the encoding matrices $k$ is a parameter which controls the rank of the encoding, akin to the number of components in dimensionality reduction techniques. The internal sizes of the dictionaries $m,t$ depend on the type of dictionaries selected discussed in the following section. 





If unconstrained, the encoding matrices $Y$ and $W$ from Eq.~\ref{obj_2} will overfit noise in the input. This may be especially exacerbated when the adopted dictionaries $\Psi$ and $\Phi$ are large and exhaustive. To alleviate this limitation, we impose sparsity on the encoding matrices akin to sparse coding techniques. In addition, the basic encoding from from Eq.~\ref{obj_2} cannot handle missing values in the input signal $X$ or irregular temporal sub-sampling. To this end, we include a mask matrix $\Omega$ ensuring a fit to observed values. 

\begin{definition}~\label{obj_missing_complete}
{\bf Sparse temporal graph signal decomposition with missing values: } \it Given a temporal graph signal $X$ with missing values, a binary mask $\Omega$ of the same size, graph $\Psi$ and time $\Phi$ dictionary matrices and rank $k$, find the encoding matrices $Y$ and $W$ which minimize the following objective:\vsb
 \begin{equation*}
\begin{aligned}
    &\underset{ Y, W } {\mathrm{argmin}} \hspace{0.1cm} \left \| \Omega \odot(X-\Psi Y W\Phi)\right\|_F^2 +\lambda_1\left\|Y
    \right\|_1 +\lambda_2\left\|W \right\|_1, 
    \end{aligned}\vsb
\end{equation*}
where $\odot$ denotes the element-wise product and $\lambda_1$ and $\lambda_2$ are sparsity regularization parameters.
\end{definition}\vsb

In the absence of missing values, we can omit the mask matrix $\Omega$ in the fit term, however, we will discuss solutions for this more general version of the objective. The sparse TGSD problem can be viewed as a generalization of matrix completion when missing values are present, and sparse coding when the input is complete. In particular, the missing value objective is a generalization of low-rank matrix 
completion~\cite{zheng2012_cvpr,Cabral2013_iccv}, since choosing trivial identity matrix dictionaries of appropriate sizes reduces our objective to that in matrix completion $\left \| \Omega\odot \left ( D-AB \right ) \right \|_F^2$. Different from the above, \ourmeth can harness the representative power of various dictionaries to capture structures in both rows and columns. 
Similarly, selecting identity matrix dictionaries in the absence of missing values reduces our objective to matrix factorization with sparsity regularization. It is important to note that our optimization solutions discussed next are applicable to any dictionaries, making TGSD general and flexible to both existing as well as "custom" new dictionaries.


\vsb
\section{Optimization solutions for TGSD}
We derive an optimization technique for the missing values objective from Def~\ref{obj_missing_complete} and discuss how it can be customized for decomposition without missing values. Since the objective from Def.~\ref{obj_missing_complete} is jointly convex, we employ Alternating Direction Method of Multipliers (ADMM)~\cite{Boyd2011ADMM} to solve it. We first introduce intermediate variables $D=X$, $Z = Y$ and $V=W$ which help ensure that all subproblems have a closed-form solution, and rewrite the objective as:
\begin{equation}
\footnotesize
\begin{aligned}
    \underset{ D,Y, W, Z, V } {\mathrm{argmin}} \hspace{0.1cm} \left \|D-\Psi Y W\Phi\right\|_F^2 +\lambda_1\left\|Z
    \right\|_1 +\lambda_2\left\|V \right\|_1 +
    & \lambda_3\left\| \Omega \odot (D-X) \right\|_F^2  \\ \hspace{01.cm}s.t.\hspace{0.1cm}
     Y=Z, W=V,  D=X
    \end{aligned}
\end{equation}\label{final_obj_01}

With some algebraic transformations, the corresponding Lagrangian function is as follows:
   \begin{equation}
   \footnotesize
\begin{aligned}
   &\mathcal{L}\ = 
    \left \|  D-\Psi Y W\Phi \right\|_F^2 +\lambda_1\left\|Z \right\|_1
    +\lambda_2\left\|V \right\|_1
    +\lambda_3\left\| \Omega \odot (D-X) \right\|_2\\
    &+ \frac{\rho_1}{2}\left\|Z - Y 
    + \frac{\Gamma_1}{\rho_1} \right\|_F^2+\frac{\rho_2}{2}\left\|V - W + \frac{\Gamma_2}{\rho_2} \right\|_F^2, 
    \end{aligned}
\end{equation}
where $\Gamma_1$ and $\Gamma_2$ are the Lagrangian multipliers and $\rho_1$ and $\rho_2$ are penalty parameters. 
Next we derive the individual variable updates for ADMM.

{\noindent \bf Update $D$:} 
Let $P= \Psi Y W \Phi$, we have the optimization problem for $Y$ as follows:
   \begin{equation}
   \footnotesize
\begin{aligned}
    \underset{ D } {\mathrm{argmin}} \hspace{0.1cm}\left \|D-P \right\|_F^2 
+\lambda_3\left\| \Omega \odot (D-X) \right\|_F^2
\label{D_up}
    \end{aligned}
\end{equation}
By taking the gradient and equating it to zero, we have 
$ D = (P+\lambda_3\Omega \odot X)\oslash (I+\lambda_3\Omega)$, where $\oslash $ is element-wise division.


{\noindent \bf Update $Y$:} 
Let $B= W \Phi$, we then have the following optimization problem for $Y$:
   \begin{equation}
   \footnotesize
\begin{aligned}
    \underset{ Y } {\mathrm{argmin}} \hspace{0.1cm}\left \|D-\Psi Y B\right\|_F^2 
+\frac{\rho_1}{2}\left\|Z-Y + \frac{\Gamma_1}{\rho_1} \right\|_F^2
    \end{aligned}
\end{equation}
Setting the gradient with respect to $Y$ to zero, we get: 
\begin{equation} \footnotesize 
\begin{aligned}
2\Psi^T \Psi Y B B ^T+\rho_1 Y =2\Psi^T D B ^T+\rho_1 Z+\Gamma_1. \end{aligned}~\label{lagY0}
\end{equation}
\noindent\emph{$\bullet$ Case 1:} If $\Psi$ is a dictionary of orthogonal atoms, we can simplify the above as follows:
\begin{equation}
\footnotesize
\begin{aligned}
   Y=(2\Psi^T D B^T+\rho_1 Z+\Gamma_1)( 2B B ^T+\rho_1 I)^{-1}.\\
   ~\label{update_Y_orth}
       \end{aligned}
\end{equation}
\noindent\emph{$\bullet$ Case 2:} If $\Psi$ is not orthogonal, we cannot solve the problem as outlined above, and thus, develop a more general solution. Note that both $BB^T$ and $\Psi^T\Psi$ are positive semi-definite and symmetric.  
Let their eigenvalue decomposition be as follows: $\Psi^T\Psi= Q_1\Lambda_1 Q_1^T$ and $ B B ^T= Q _2\Lambda_2Q _2^T$, where $Q_1$, $Q_2$ are orthonormal eigenvector matrices and $\Lambda_1$, $\Lambda_2$ are diagonal non-negative eigenvalue matrices. Let $\Pi_1$ be the quantity on the right side of Eq.~\ref{lagY0}, and let us multiply the equation on both sides by the eigenvector matrices as follows: 
\begin{equation}
\footnotesize
\begin{aligned}
& \Pi_1=2\Psi^T D B ^T+\rho_1 Z +\Gamma_1\\
& \Rightarrow  \Pi_1=2 Q_1 \Lambda_1 Q_1^T Y Q_2\Lambda_2  Q_2^T+\rho_1 Y\\
 & \Rightarrow  Q_1^T\Pi_1 Q_2=2\Lambda_1 Q_1^T Y Q_2\Lambda_2+ \rho_1 Q_1^T Y Q_2
    \end{aligned}~\label{Q1Q2_eq}
\end{equation}
Substituting $E_1=Q_1^T Y Q_2$ in Eq.~\ref{Q1Q2_eq} we obtain $  Q_1^T\Pi_1 Q_2=2\Lambda_1E_1\Lambda_2+ \rho_1E_1$, and an element-wise solution for $E_1$ as follows:
 \begin{equation}
 \footnotesize
\begin{aligned}
  {[E_1]}_{(i,j)}= {[Q_1^T\Pi_1Q_2]}_{(i,j)}/2[\Lambda_1]_{(ii)}[\Lambda_2]_{(jj)} + \rho_1
    \end{aligned}
\end{equation}
Finally, we update $Y$ based on $E_1$: $Y=Q_1 E_1 Q_2^T$.  

{\noindent  \bf Update $W$:}
When we fix other variables and set $A= \Psi Y$, the problem w.r.t. $W$ is reduced to: 
   \begin{equation}
   \footnotesize
\begin{aligned}
    &\underset{ W } {\mathrm{argmin}} \hspace{0.1cm}\left \|D-A W\Phi\right\|_F^2 
+\frac{\rho_2}{2}\left\|V-W + \frac{\Gamma_2}{\rho_2} \right\|_F^2
\label{Up_w}
    \end{aligned}
\end{equation}

\noindent\emph{$\bullet$ Case 1:} For orthogonal $\Phi$ we can set the gradient w.r.t. $W$ to zero, obtaining:

$W =(2 A ^T A +I \rho_2 )^{-1}(2A ^T X \Phi^T +\rho_2 V+ \Gamma_2)$.

\noindent\emph{$\bullet$ Case 2:} For non-orthogonal $\Phi$, we get $W= Q_3 E_2 Q_4^T$, 
where $E_2{(i,j)}=[Q_3^T\Pi_2 Q_4]_{i,j}/2[\Lambda_4]_{ii}[\Lambda_3]_{jj}+ \rho_2$ and $(Q_3,\Lambda_3$ and $(Q_4,\Lambda_4)$ are the (eigenvector, eigenvalue) matrices of $A^TA$ and $\Phi\Phi^T$, respectively.


 
 


{\noindent \bf Update $Z$ and $V$:}
The problems w.r.t  $Z$ and $V$ are: 
\begin{equation}
\footnotesize
\begin{aligned}
 \begin{cases}
   \underset{Z}{\mathrm{argmin}} \hspace{0.1cm} \lambda_1\left \| Z \right \|_1 + \frac{\rho_1}{2} \left \| Z- Y + \frac{\Gamma_1}{\rho_1} \right \|_F^2 \\ 
    \underset{V}{\mathrm{argmin}} \hspace{0.1cm} \lambda_2\left \| V \right \|_1 + \frac{\rho_2}{2} \left \| W - V + \frac{\Gamma_2}{\rho_2} \right \|_F^2
\end{cases}
\end{aligned}
\end{equation}
Closed-form solutions are available due to~\cite{Lin2013TheAL}:
\begin{equation}
\footnotesize
\begin{aligned}
\begin{cases}
  Z_{ij} = sign\left (  H^{(1)}_{ij}\right )\times max\left ( \left | H^{(1)}_{ij}  \right | -\frac{\lambda_1}{\rho_1},0 \right ) \\ 
  V_{ij} = sign\left (  H^{(2)}_{ij}\right )\times max\left ( \left | H^{(2)}_{ij}  \right | -\frac{\lambda_2}{\rho_2},0 \right ),  
 \label{up_l1}
\end{cases} 
\end{aligned}
\end{equation}
where $H^{(1)}= Y - \frac{\Gamma_1}{\rho_1}$ and $H^{(2)}= W - \frac{\Gamma_2}{\rho_2}$.

\eat{
{\noindent \bf Graph signal dictionary~\cite{dong2019learning}:}
The eigenvectors of the Laplacain matrix constitute the columns of the GFT. 
\todo{need to add more details}

{\noindent \bf Graph-Haar Wavelet:}
The Graph-Haar Wavelet was formulated by the authors of \cite{findingGems2019}. The authors use an approximation of the Fiedler vector to perform clustering of the graph. The Fideler vector separates the nodes in a graph into two sets. One set has positive values and the other has negative.  These sets can be used to cluster the nodes in the graph. These sets can be clustered in a recursive manner until signal nodes or a constant polarity vector is obtained.
Let bisection for the $\iota_{th}$ fidler vector ($v^\iota_f$)
be represented by.

\begin{equation}
\begin{aligned}
\Omega^\iota_1={i|v^\iota_f[i]\geq 0}\\
\Omega^\iota_2={i|v^\iota_f[i]\geq0}
      \end{aligned}
\end{equation}

Once this the authors have this  clustering they then transform this information into a orthonormal Haar-like wavelet basis. Explicitly, the first function is constant over the graph. 

\begin{equation}
\begin{aligned}
\phi_0[i]=\frac{1}{\sqrt{N}} \forall(i),
      \end{aligned}
\end{equation}

\begin{equation}
\begin{aligned}
\phi_\iota[i]= 
\begin{cases}
\frac{\sqrt{|\Omega^
\iota_2|}}{\sqrt{|\Omega^
\iota_1|}\sqrt{|\Omega^
\iota_1| + |\Omega^
\iota_2|} } i  \in \Omega^\iota_1,
\\
-\frac{\sqrt{|\Omega^
\iota_1|}}{\sqrt{|\Omega^
\iota_2|}\sqrt{|\Omega^
\iota_1| + |\Omega^
\iota_2|} } i  \in \Omega^\iota_2,
 \\ 
 0& \text{else }
\end{cases}
\end{aligned}
\end{equation}

The accumulated set of basis function ${\phi_\iota}_\iota$ constitues the columns of the Graph-Haar base dictionary.

{\noindent \bf  Discrete Fourier Transform}

A fourier transform is a widely adopted way of decomposing function into the frequencies that make it. It is commonly defined as 

\begin{equation}
    \~{f}(\xi)=\int_{-\infty}^{\infty} f(x)e^{-2\pi i x \xi}dx
\end{equation}

For any real number $\xi$. Using the principles of this we can construct a DFT matrix. 
Let $\omega=e^{\frac{-2\pi i}{ N}} $. We can construct a DFT as follows.

\begin{equation}
\footnotesize
W=
\begin{bmatrix}
1 & 1 & 1 & 1 & ... & 1\\
1 & \omega & \omega^2 & \omega^3 & ... & \omega^{N-1}\\
1 & \omega^2 & \omega^4 & \omega^6 & ... & \omega^{2(N-1)}\\
1 & \omega^3 & \omega^6 & \omega^9 & ... & \omega^{3(N-1)}\\
\vdots & \vdots & \vdots & \vdots & \ddots & \vdots & \\
1 & \omega^{(N-1)} & \omega^{2(N-1)} & \omega^{3(N-1)} & ... & \omega^{(N-1)(N-1)}\\

\end{bmatrix}
\end{equation}

If we normalize by multiplying by  $\frac{1}{\sqrt{N}}$ the matrix becomes orthonormal which brings addtional benefit.

{\noindent \bf  Ramanujan periodic dictionary~\cite{tennetiTSP2015}:}
Ramanujan periodic dictionary 
has been proposed by Tenneti~\etal
for learning the periods in
time series~\cite{tennetiTSP2015}. 
Ramanujan periodic dictionary ${R}$ is written
as ${R} = \left [ {\Phi}_1 ,..,  {\Phi}_{g_{max}}   \right ]$, where $g_{max}$ is the maximum period.
The subdictionary ${\Phi}_i$ represents the periodic basis of period $g_{i}$. Please refer to ~\cite{tennetiTSP2015} for more details.
With this dictionary, we can represent the periodic time series as the linear combination of the basis in it.

{\noindent \bf Spline dictionary~\cite{Goepp2018Spline}:}
B-splines is often used to fit smooth curves~\cite{Goepp2018Spline,Eilers2010SKP}, therefore, one can build a dictionary upon it,
where input time series can be approximated as the linear combination of the basis in the dictionary.
More specifically,
we construct the spline from B-splines basis functions $B_{i,d}(u)$ which can be defined recursively by the Cox-de-Boor formula:
$B_{i,0} = 1$  $\text{if}\ u_i \leq u < u_{i+1}$, $0$ otherwise.
\begin{equation*}
\footnotesize
B_{i,p} = \frac{u-u_i}{u_{i+p}-u_i}B_{i,p-1}(u) + \frac{u_{i+p+1}-u}{u_{i+p+1}-u_{i+1}}B_{i+1,p-1}(u)
\end{equation*}
Each $B_{i,d}(u)$ is nonzero on the range of $[u_i, u_{i+d+1})$. 

}


\begin{center}
	\begin{algorithm}[!t]
		\footnotesize
		\caption{\ourmeth (with missing values)} 
		\label{alg:opt}
		\begin{algorithmic}[1]
			\Require{ Input $X$, mask $\Omega$, dictionaries $\{\Psi, \Phi\}$, $k$, $\lambda_1,\lambda_2$ 
			}
 \State  Initialize $Y=Z=\textbf{1}$, $W=V=\textbf{1}$ 
 
                \While{ not converged } 
                    
                    \State $P=\Psi Y W \Phi$ 
                    \State  $D = (P+\lambda_3\Omega \odot X)\oslash (I+\lambda_3\Omega)$
                     
                     \State $B=W \Phi$
                    \State  $Y=(2\Psi^T D B^T+\rho_1 Z+\Gamma_1)( 2B B ^T+\rho_1 I)^{-1}$
                    
                     \State $   A=\Psi Y$
                     \State   $W =(2 A ^T A +I \rho_2 )^{-1}(2A ^T X \Phi^T +\rho_2 V+ \Gamma_2)$
                     
                     \State  $  V_{ij} = sign\left (  H_{ij}\right )\times max\left ( \left | H_{ij}  \right | -\frac{\lambda_1}{\rho_2},0 \right )  $
                     \State   $Z_{ij} = sign\left (  H_{ij}\right )\times max\left ( \left | H_{ij}  \right | -\frac{\lambda_1}{\rho_1},0 \right )  $
                     
                      \State $\Gamma_1^{i+1} =\Gamma_1^{i} +\rho_1\left ( Z-Y  \right ) $
                      \State $\Gamma_2^{i+1}=\Gamma_2^{i} + \rho_2\left ( V-W  \right )$
                      \State $i\leftarrow i+1$ 
                \State Convergence condition: $\left | f^{i+1}-f^{i} \right | \leq \varepsilon $, where $f^{i+1}$ and $f^{i}$ are the objective values of Eq.~\ref{final_obj_01} at iterations $i+1$ and $i$.
                
                \EndWhile        
		\end{algorithmic}
	\end{algorithm}
\end{center}

\noindent{\bf The overall TGSD algorithm.}
We show all updates within the overall optimization procedure in  Alg.~\ref{alg:opt}. We repeat updates from Step $3$ to Step $12$ until convergence.
We demonstrate experimentally that key hyper-parameters like the number of components $k$ can be in a supervised manner by cross-validation. A similar approach can be employed for the sparsity regularizers $\lambda_1$ and $\lambda_2$.  

The complexity of \ourmeth is dominated by the matrix inversions in Step $6$ and $8$. Although the complexity of a quadratic matrix inversion is in general cubic, in practice our overall running time is practical due to fast convergence and the ability to work with reduced dictionaries as demonstrated in scalability experiments in Fig.~6. 
The optimization procedure can be minimally altered for the case without missing values by setting $\Omega$ to an all-ones matrix and removing the optimization of $D$ from step 4, in which the overall complexity remains the same. 
For orthogonal dictionaries we can use the more efficient updates from step~8 and step~6 for $W$ and $Y$, respectively.
When either of the dictionaries (graph or time) is non-orthogonal, we need to work with the general solutions employing eigenvalue decomposition (Case 2 in the updates of $Y$ and $W$). 
Non-orthogonal versions come with extra cost due to the eigendecompositions, 
however the overall complexity remains unchanged since the inversions remain the most costly steps.



\section{Dictionaries for TGSD}
Our framework can flexibly accommodate many graph and time dictionaries. For the purposes of evaluation we employ several popular alternatives listed in Tbl.~\ref{table:dictionary}. 
We experiment with multiple versions of TGSD employing different combinations of graph and time dictionaries. Our naming convention specifies them in order. For example, when employing GFT for $\Psi$ and DFT for $\Phi$, we denote our method: \ourmeth G+D. We next provide a brief definition of the dictionaries and refer to relevant work for more details.

The {\bf Graph Fourier
Transform (GFT)~\cite{dong2019learning}} basis consists of the eigenvectors $U$ of the graph Laplacian matrix $L$, where $L = U\Lambda U^T$. Graph signal processing draws a parallel between GFT and the discrete Fourier transform (DFT) where small eigenvalues in $\Lambda$ 
correspond to ``low-frequency'' components as they tend to identify larger regions in the graph structure. These "low-frequency" can also be used capture higher order dependencies between nodes while the "high-frequencies" capture more local dependencies~\cite{ortega2018graph}.  
GFT is orthonormal since $U$ is an eigenvector matrix.

{\bf Graph-Haar Wavelets~\cite{crovella2003graph}} have been adopted for many graph data analytics tasks~\cite{crovella2003graph} and are central to one of our baselines Gems-HD~\cite{findingGems2019}. An orthonormal basis is computed by thresholding the Fiedler vector, obtaining recursive bisections of the graph. Let $V'$ be a subset of nodes obtained in the recursive partitioning tree and $V'_1$ and $V'_2$ the two subsets obtained by thresholding the Fiedler vector at $0$ for the subgraph induced by $V'$. The basis function $\phi'(v)$ for $V'$ is defined as:
\begin{equation}
\footnotesize
\begin{aligned}
\phi'(v)= 
\begin{cases}
\frac{\sqrt{|V'_2|}}{\sqrt{|V'_1|}\sqrt{|V'_1| + |V'_2|} } &\text{ if } v  \in V'_1,
\\
-\frac{\sqrt{|V'_1|}}{\sqrt{|V'_2|}\sqrt{|V'_1| + |V'_2|} } &\text{ if } v  \in V'_2,
\\
 0& \text{ if } v \notin V 
\end{cases}
\end{aligned}
\end{equation}

\begin{table}[!t]
\footnotesize
\setlength\tabcolsep{3 pt}
\centering
 \begin{tabular}{c| c| c ||c| c |c |} 
  & \multicolumn{2}{c||}{Graph dictionaries} &  \multicolumn{3}{c|}{Temporal dictionaries} \\ 
 \hline
  & GFT (G) & Wavelet (W)&  DFT (D) & Ramanujan (R) & Spline (S) \\ [0.5ex] 
 \hline
 Orthogonal & \checkmark & \checkmark & \checkmark  & &\\
 \hline
 Parameter-free  & \checkmark & \checkmark & \checkmark & &\\
 \hline
\end{tabular}
 \caption{\footnotesize Summary of dictionaries we experiment with.}
\label{table:dictionary}\vsa\vsa\vsb
\end{table}
The {\bf Discrete Fourier Transform (DFT)~\cite{tennetiTSP2015}} dictionary $W$ for temporal signals of length $N$ is defined as: 
\begin{equation}
\footnotesize
W= \frac{1}{\sqrt{N}}
\begin{bmatrix}
1 & 1 & 1 & ... & 1\\
1 & \omega & \omega^2 &  ... & \omega^{N-1}\\
\vdots & \vdots & \vdots & \ddots & \vdots & \\
1 & \omega^{(N-1)} & \omega^{2(N-1)} & ... & \omega^{(N-1)(N-1)}\\
\end{bmatrix},
\end{equation}
where $\omega=e^{\frac{-2\pi i}{ N}}$ and $i$ is the imaginary unit. This basis is unitary.




The {\bf Ramanujan periodic dictionary~\cite{tennetiTSP2015}}, similar to DFT, is applicable to periodic signals and is constructed by stacking period-specific sub-matrices of varying width
${R} = \left [ {\Phi}_1 ,..,  {\Phi}_{g_{max}}  \right ]$, where $g_{max}$ is the a maximum modeled period and 
${\Phi}_i$ is the periodic basis of period $g_{i}$. 
Period-specific matrices ${\Phi}_g  = \left [ {D}_{d_1}, {D}_{d_2},...{D}_{d_K} \right ]$ have columns determined by the divisors $\left \{ d_1,d_2,...d_K \right \}$ of $g$.   
${D}_{d_i}\in\mathbb{R}^{g\times \phi\left ( d_i \right ) }$ is a periodic basis for period $d_i$ of the following circulant matrix form:
\begin{equation}
\footnotesize
    { D}_{d_i} = \begin{bmatrix}
C_{d_i}(0) & C_{d_i}(g-1) & ... &C_{d_i}(1) \\ 
 C_{d_i}(1)&C_{d_i}(0)  & ... &C_{d_i}(2) \\ 
 ...& ... & ... & ...\\ 
C_{d_i}(g-1) & C_{d_i}(g-2) &  ...&C_{d_i}(0) 
\end{bmatrix},
\end{equation}
where the number of columns, $\phi\left ( d_i \right )$ denotes the Euler totient function. Elements $C_{d_i}(g)$ are computed as the Ramanujan sum:
\begin{equation}
\footnotesize
    C_{d_i}(g) = \sum_{k=1,gcd(k,{d_i})=1}^{d_i} e^{j2\pi kg/{d_i}},
\end{equation}
where $gcd(k,d_i)$ is the greatest common divisor of $k$ and ${d_i}$. This dictionary is not orthogonal.


{The \bf Spline dictionary~\cite{Goepp2018Spline}} is applicable for encoding smoothly-evolving time series and can be constructed by employing B-splines $B_{i,d}(u)$, defined by the Cox-de-Boor formula:
\begin{equation*}
\footnotesize
B_{i,p} = \frac{u-u_i}{u_{i+p}-u_i}B_{i,p-1}(u) + \frac{u_{i+p+1}-u}{u_{i+p+1}-u_{i+1}}B_{i+1,p-1}(u),
\end{equation*}
where $B_{i,0} = 1$  $\text{if}\ u_i \leq u < u_{i+1}$, and $0$ otherwise. $B_{i,d}(u)$ is non-zero in the range of $[u_i, u_{i+d+1})$. 
This dictionary is non-orthogonal.


\section{Experimental evaluation}
\label{sec:exp}

\begin{table}[!t]
\footnotesize
\setlength\tabcolsep{3 pt}
\centering
 \begin{tabular}{|l|c|c|} 
 \hline
  {\bf Task} & {\bf Baselines} & {\bf Quality metric} \\ 
 \hline
1. Decomposition & MCG, LRDS, GEMS-HD, SVD & RMSE v.s. model size \\
 \hline
2. Imputation & MCG, LRDS, GEMS-HD, BRITS & RMSE v.s. \% missing \\
 \hline
3. Interpolation & MCG, LRDS, GEMS-HD, BRITS & RMSE v.s. \% missing \\
  \hline
4. Node clustering & CCTN, PCA & Accuracy \\
   \hline
5. Period detection & NPM, FFT, AUTO & Accuracy \\ \hline
\end{tabular}
 \caption{\footnotesize Summary of evaluation tasks, baselines and metrics.}
\label{table:tasks}\vsa\vsa\vsb
\end{table}

We evaluate \ourmeth on five tasks listed in Tbl.~\ref{table:tasks} and compare against a total of $10$ baselines across tasks (detailed in Sec.~\ref{sec:baselines}) on $6$ datasets (Tbl.~\ref{table:datasets}). Our goal is to test the accuracy, scalabilty and conciseness of our model. 
We quantify model conciseness in terms of number of non-zero  values (nnz) in its representation and measure scalability in terms of running time of single-core MATLAB implementations. \vsa

\vsa
\subsection{Datasets.} Table~\ref{table:datasets} shows the statistics of our real-world and synthetic datasets.  
The graph structure in our {\bf Synthetic} data consists of $7$ overlapping groups (on average $10\%$ of nodes in each group belong to other groups).
We generate independently periodic time series for each group similar to the protocol in~\cite{tennetiTSP2015}. We scale node signals randomly (uniform in $[1,10]$) and add Gaussian noise at $SNR=10$. 

We also employ $2$ real-world data-sets for reconstruction, imputation and interpolation experiments; and $3$ more real-world data sets with ground-truth communities to evaluate clustering. 
The {\bf Bike}~\cite{BostonBike} dataset contains daily bike check-out counts at rental stations in Boston. Pairs of stations are connected by an edge if within approximately 2.22 km. 
The graph in the {\bf Traffic}~\cite{LA_traffic} dataset corresponds to a highway network where nodes are locations of inductive loop sensors. We use the average speed (at a resolution of 3 hours) at sensors as our evolving graph signal. 
We normalize both the Bike and Traffic datasets by using the MATLAB's function \emph{normr} which scales each to a norm of $1$: $\left\|X_{i}\right\|_2=1, \forall i \in [1,n]$.

The {\bf Reality Mining}~\cite{eagle2006reality} tracks the number of hourly interactions of $142$ people at MIT where an edge between two individuals exists if they interacted at least $50$ times. 
We employ lab group membership (provided in the dataset) as community ground truth.
The two {\bf Reddit} datasets~\cite{zhang2019PERCeIDs} are derived from public reddit comments between $2008$ and $2015$. Undirected user-user edges exist if a user replied at least once to another user's top-level post. {\bf Reddit-epi} consists of $242$ users who posted in one of $25$ subredits dedicated to popular shows. {\bf Reddit-sp} involves $625$ users in $6$ sports-related subreddits. Resolution of the datasets is hourly in both cases. The subreddit in which a user participates the most is treated as the ground truth community assignment for that user. 




\vsb
\subsection{Baselines}
\label{sec:baselines} In the decomposition, imputation and interpolation tasks, we employ three baselines: \textbf{MCG}~\cite{kalofolias2014matrix}, \textbf{LRDS}~\cite{timeVaryingGraph} and \textbf{Gems-HD}~\cite{findingGems2019} (Tbl.~\ref{table:tasks}). \textbf{MCG} imputes matrix missing values based on rank minimization and by incorporating a row and a column similarity graph. We define the column graph by connecting neighboring time-steps, thus enforcing smoothness in time. \textbf{LRDS} also employs rank minimization within a regularizer that combines graph and temporal smoothness. \textbf{Gems-HD} is a state-of-the-art graph signal processing method for evolving graph signals. As it does not handle missing values, we impute them as a pre-processing step employing~\cite{moritz2017imputets} and denote the resulting 2-step method \textbf{Gems-HD+}. We also compare to \textbf{BRITS}~\cite{WeiNIPS2018} which interpolates missing values in times series by employing a bi-directional recurrent neural network. It also takes advantage of the correlation structure among uni-variates. \emph{We perform exhaustive hyper-parameter search for all baselines (details in the supplement) and report results for the best parameter settings.}


\begin{table}[t]
\footnotesize
\setlength\tabcolsep{3 pt}
\centering
 \begin{tabular}{|c| c| c| c| c |c |} 
 \hline
 {\bf Dataset} & {\bf Nodes} & {\bf Edges} & {\bf t} & {\bf k} & {\bf Resolution} \\
 \hline
    Synthetic &  175-50k  & 2k-500k  & 200-50k &  7 & - \\ 
 \hline
Bike~\cite{BostonBike} &  142  & 3446  &  628 & NA & 1 day  \\
 \hline
 Traffic~\cite{LA_traffic}  & 1938 & 5318 &  720 & NA & 3 hour  \\ 
 \hline
 Reality Mining~\cite{eagle2006reality} & 94 & 1546 & 8636 & 5 & 1 hour  \\
  \hline
  Reddit-epi~\cite{zhang2019PERCeIDs} & 242 & 1220 & 3728 & 25 & 1 hour  \\
\hline
Reddit-sp~\cite{zhang2019PERCeIDs} & 625 & 2872 & 4325 & 6 & 1 hour 
\\
\hline
\end{tabular} 
 \caption{\footnotesize Summary of evaluation datasets.}
\label{table:datasets}\vsa\vsa\vsb
\end{table}

\begin{figure} [!t]
    \centering
     \includegraphics[width=0.46\textwidth,trim={1.5cm 0cm 3.2cm 1cm},clip]{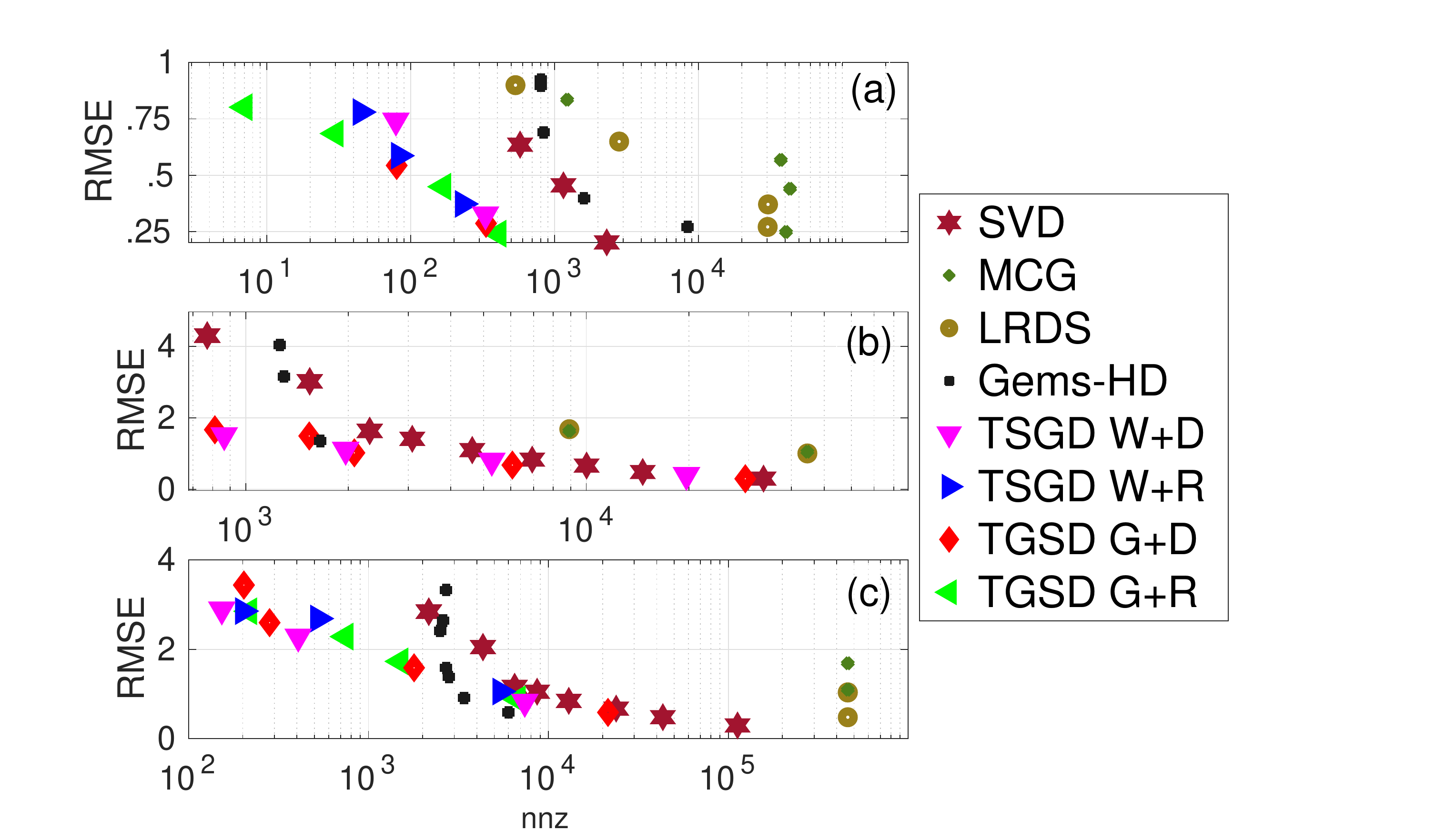}
        \label{fig:syn_nnz}
\vsa\vsb
    \caption{\footnotesize Decomposition quality as a function of model size. (a) Synthetic, (b) Bike and (c) is Traffic.}\vsa\vsa
    \label{fig:Coefficient_vs_reconstruction}
\end{figure}

For clustering we compare against the state-of-the-art for graph time series clustering \textbf{CCTN}~\cite{liu2019coupled}, and to \textbf{PCA}~\cite{jolliffe2016principal}. We employ k-means to cluster the low-dimensional representations of competing methods, following the protocol from \textbf{CCTN}~\cite{liu2019coupled}.
We employ three baselines for period detection: the state-of-art period learning method \textbf{NPM}~\cite{tennetiTSP2015}, \textbf{FFT}~\cite{li2010mining}, and a method combining auto-correlation and Fourier transform \textbf{AUTO}~\cite{LiWH15TKDE}. When the input contains missing values, we first impute using splines~\cite{moritz2017imputets} resulting in baselines: \textbf{NPM+}, \textbf{FFT+} and \textbf{AUTO+}.














\vsb
\subsection{Graph signal decomposition} \label{text:graph signal decomposition}
We first evaluate the ability of our model to succinctly reconstruct evolving graph signals. We vary the parameters of all competing methods and report the RMSE of their reconstruction as a function of the number of non-zero model coefficients (NNZ) in Fig.~\ref{fig:Coefficient_vs_reconstruction}. We plot only Pareto-optimal points for all methods, i.e. parameter settings resulting in dominated points are omitted for clarity of the figures. 
We add SVD in the comparison as a ``strawman'' baseline. 

The variants of \ourmeth dominate all baselines in the small nnz range, since the dual encoding requires fewer coefficients to represents trends in the signal. 
Since our synthetic datasets have a clear community structure with periodic and synchronous within-community signals, \ourmeth is able to represent the data using a small number of coefficients (Fig.~\ref{fig:Coefficient_vs_reconstruction}(a)). \ourmeth's quality similarly dominates alternative in real-world datasets for small nnz. 

Gems-HD achieves a good-quality reconstruction when larger models are allowed in the Traffic dataset (Fig.~\ref{fig:Coefficient_vs_reconstruction}(c)) as it employs a small number of the atoms to express a signal on a graph. These atoms select the communities in the graph and a coefficient matrix is used to reconstruct community signals. With sufficient number of coefficients Gems-HD is able to express a large portion of the communities and their conserved temporal patterns in the Traffic dataset. 
LRDS and MCG require significantly more coefficients on all datasets (the x-axis is logarithmic), as their primary goal is data imputation and their models are not explicitly sparsified.  
\ourmeth outperforms SVD following a similar trend with increasing models size. The quality advantage of \ourmeth is thanks to the structural knowledge encoded in the dictionaries enabling sparser encodings. 

\begin{figure} [!t]
    \centering
    \subfigure [Synthetic-impute 
    ]
    {
        \includegraphics[width=0.165\textwidth]{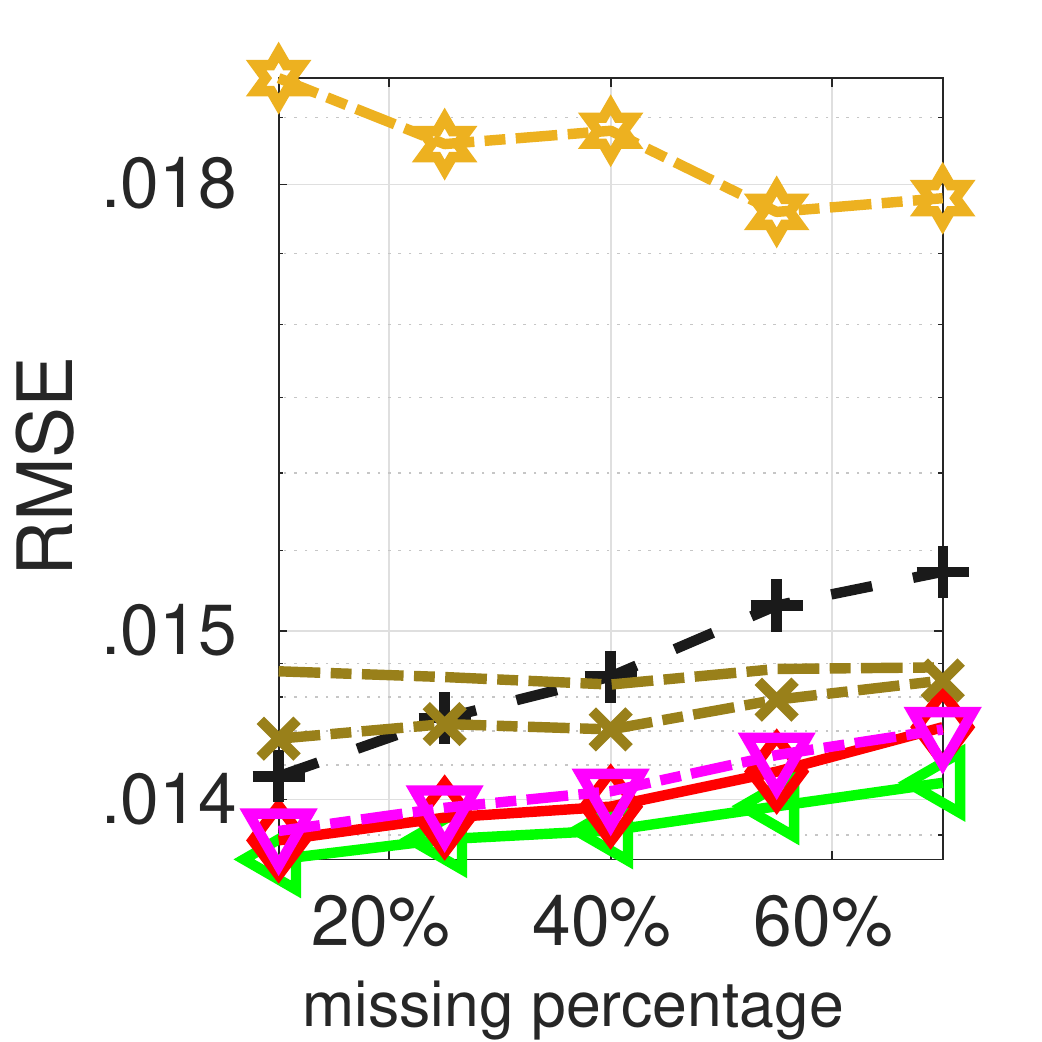}
        \label{fig:syn_random}
    }\hspace{-0.1in}
    \subfigure [ Synthetic-interpolate 
    ]
    {
        \includegraphics[width=0.17\textwidth]{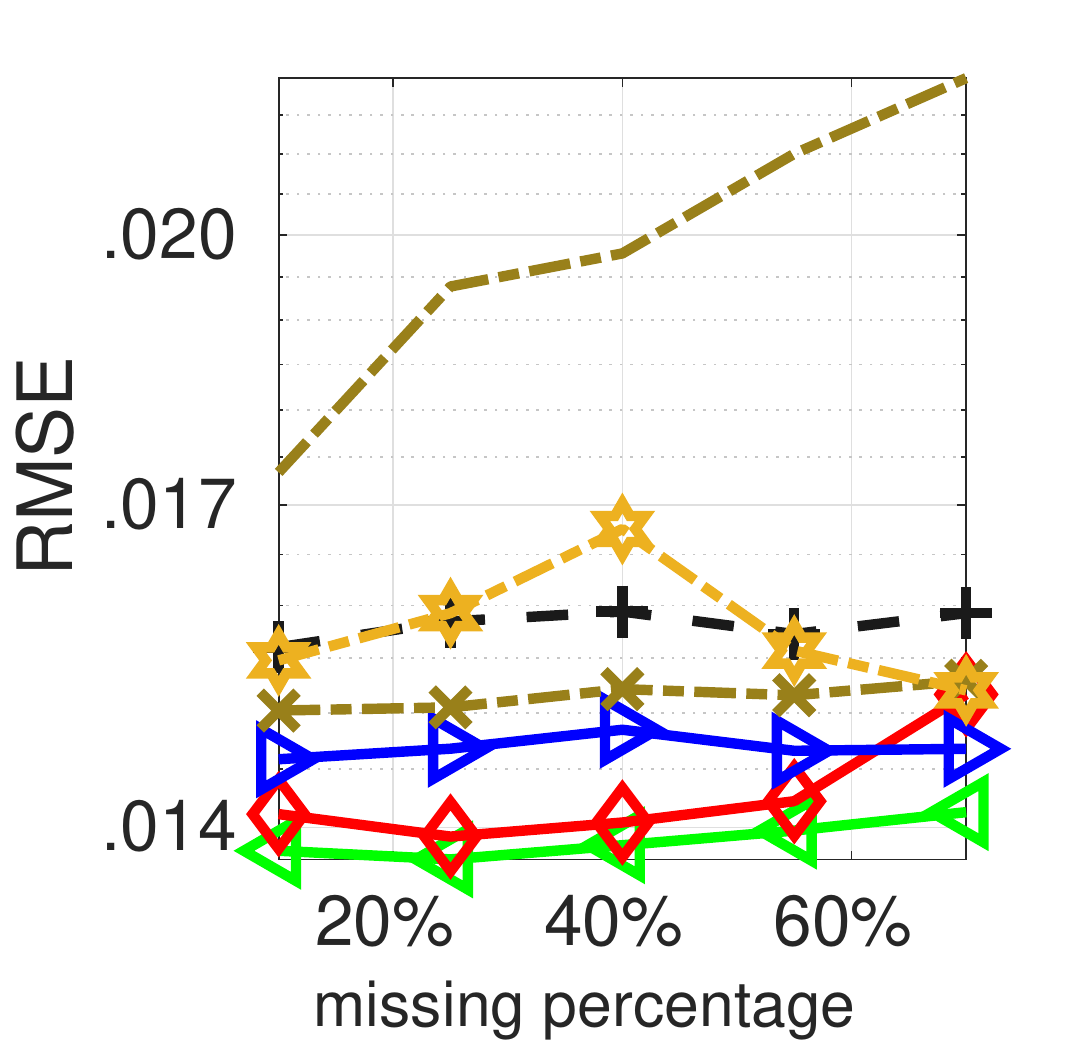}
        \label{fig:syn_col}
    }
     \subfigure[Synthetic-legend]
      {
        \includegraphics[trim={0 -0.3cm 1 1 },width=0.105\textwidth]{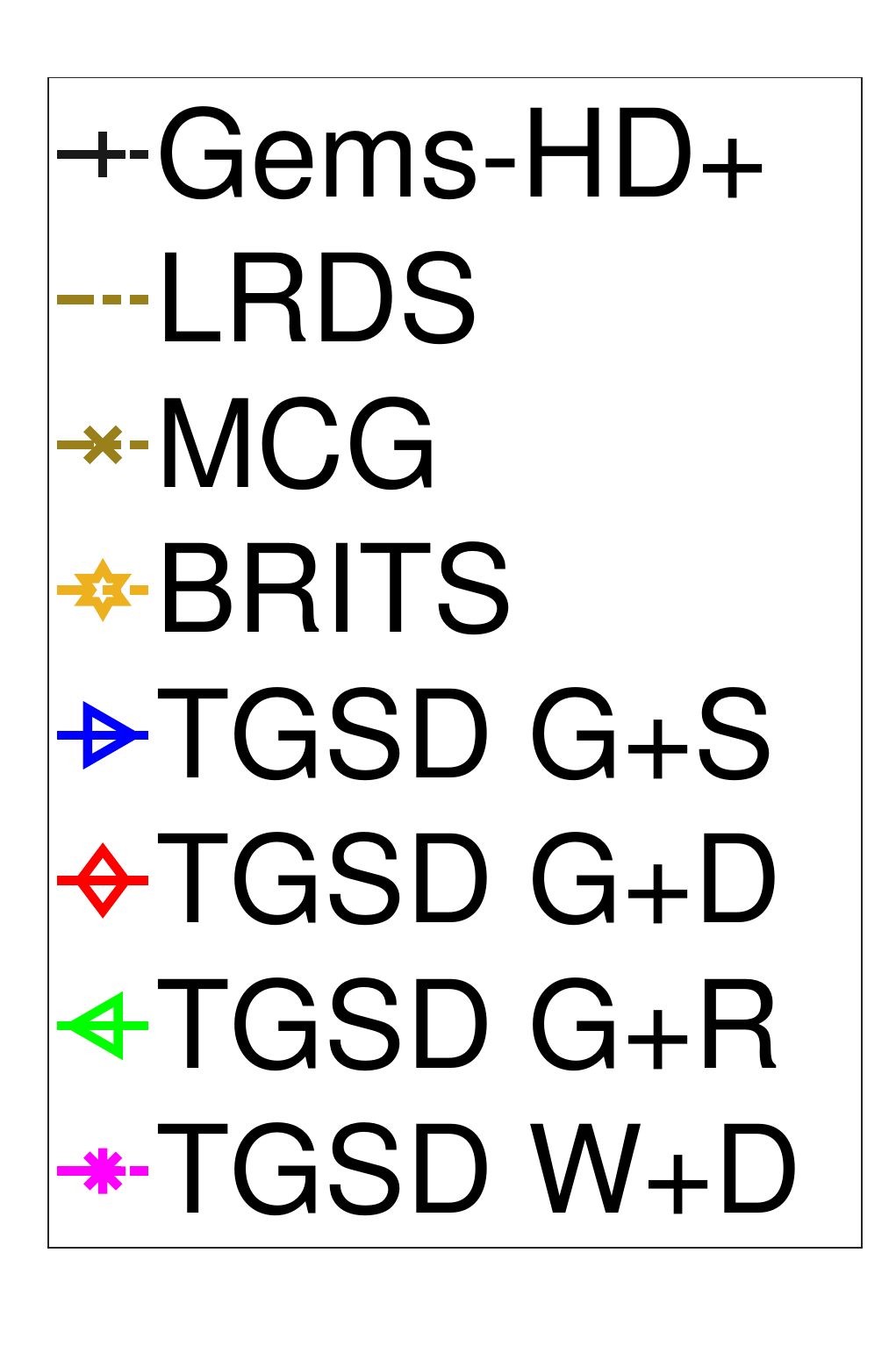}
        \label{fig:syn_legend}
    }\hspace{-0.1in}
        \subfigure [Bike-impute 
        ]
    {
        \includegraphics[width=0.17\textwidth]{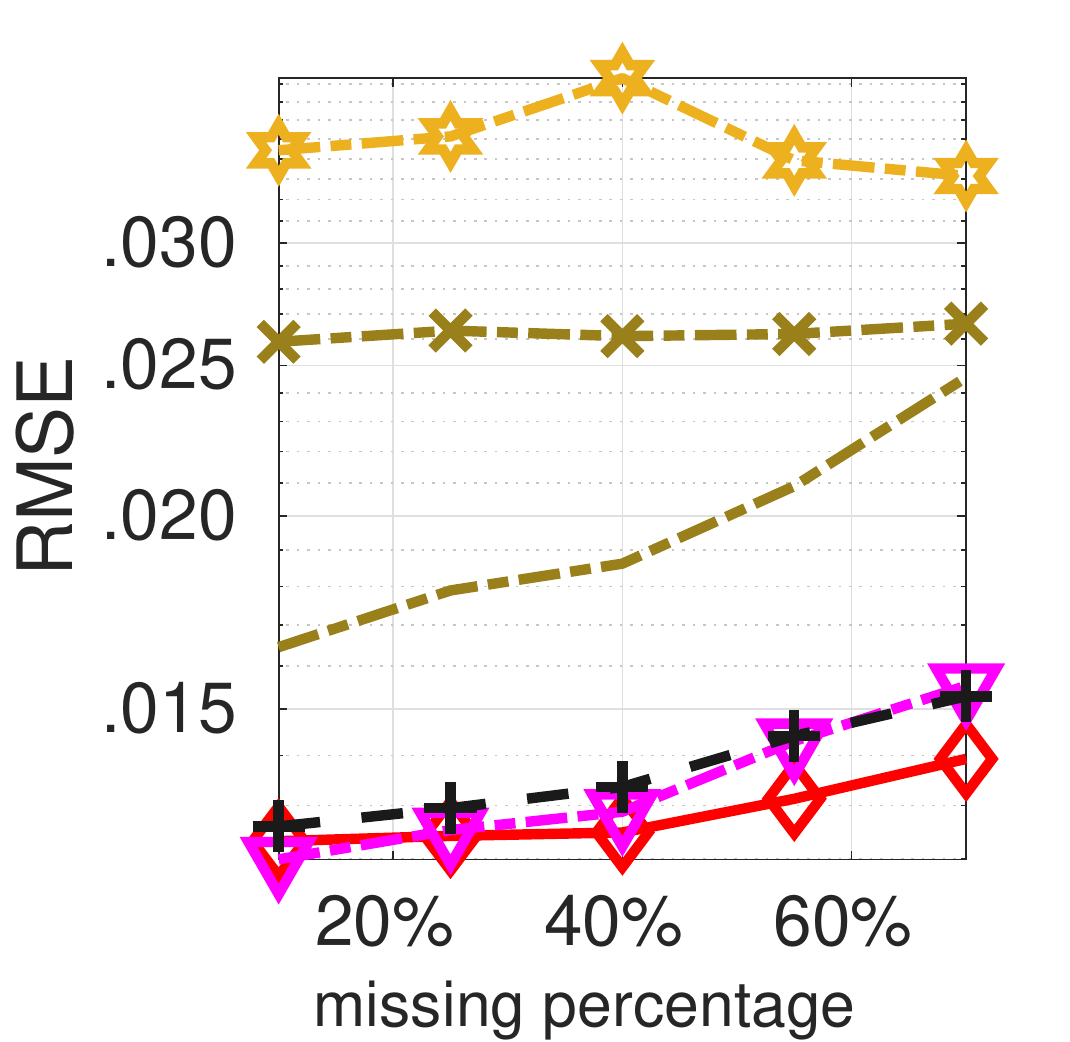}
        \label{fig:bike_random}
    }\hspace{-0.1in}
         \subfigure [ Bike-interpolate  
        ]
    {
        \includegraphics[width=0.165\textwidth]{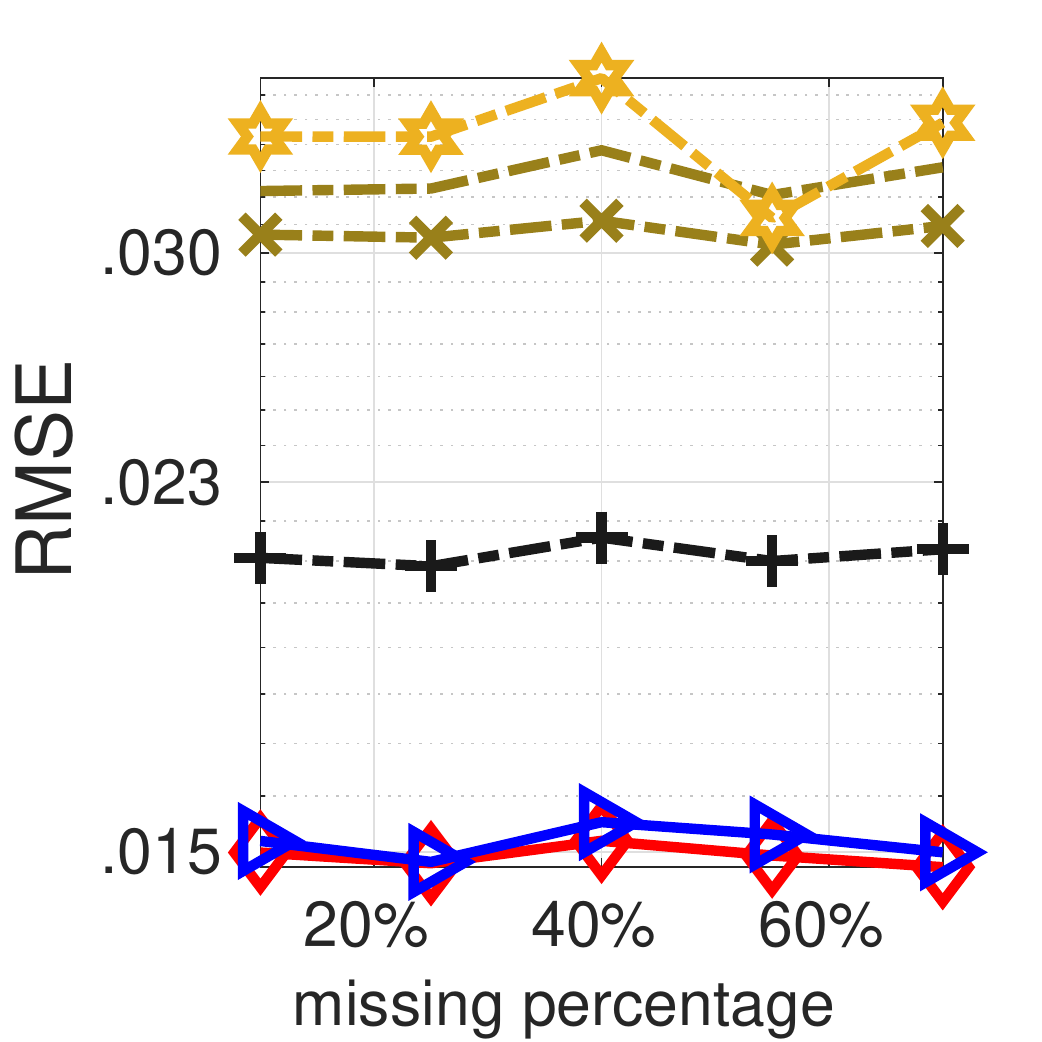}
        \label{fig:boston_col}
            }\hspace{-0.1in}
         \subfigure [ Bike-legend  
        ]
    {
        \includegraphics[trim={0 -2.3cm 1 1 },width=0.11\textwidth]{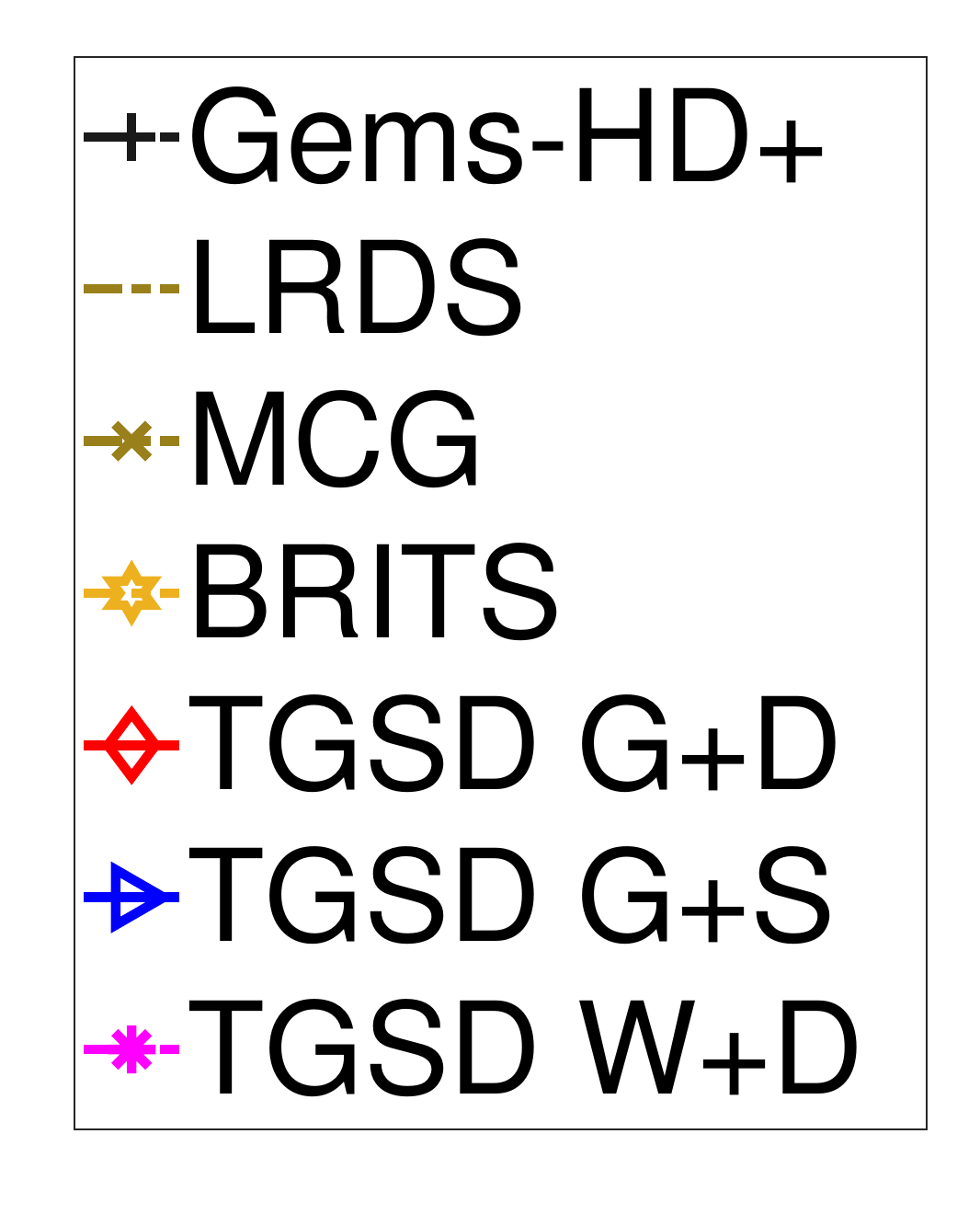}
        \label{fig:boston_legend}
            }\hspace{-0.1in}
            \subfigure [  Traffic-impute 
           ]
    {
        \includegraphics[width=0.17\textwidth]{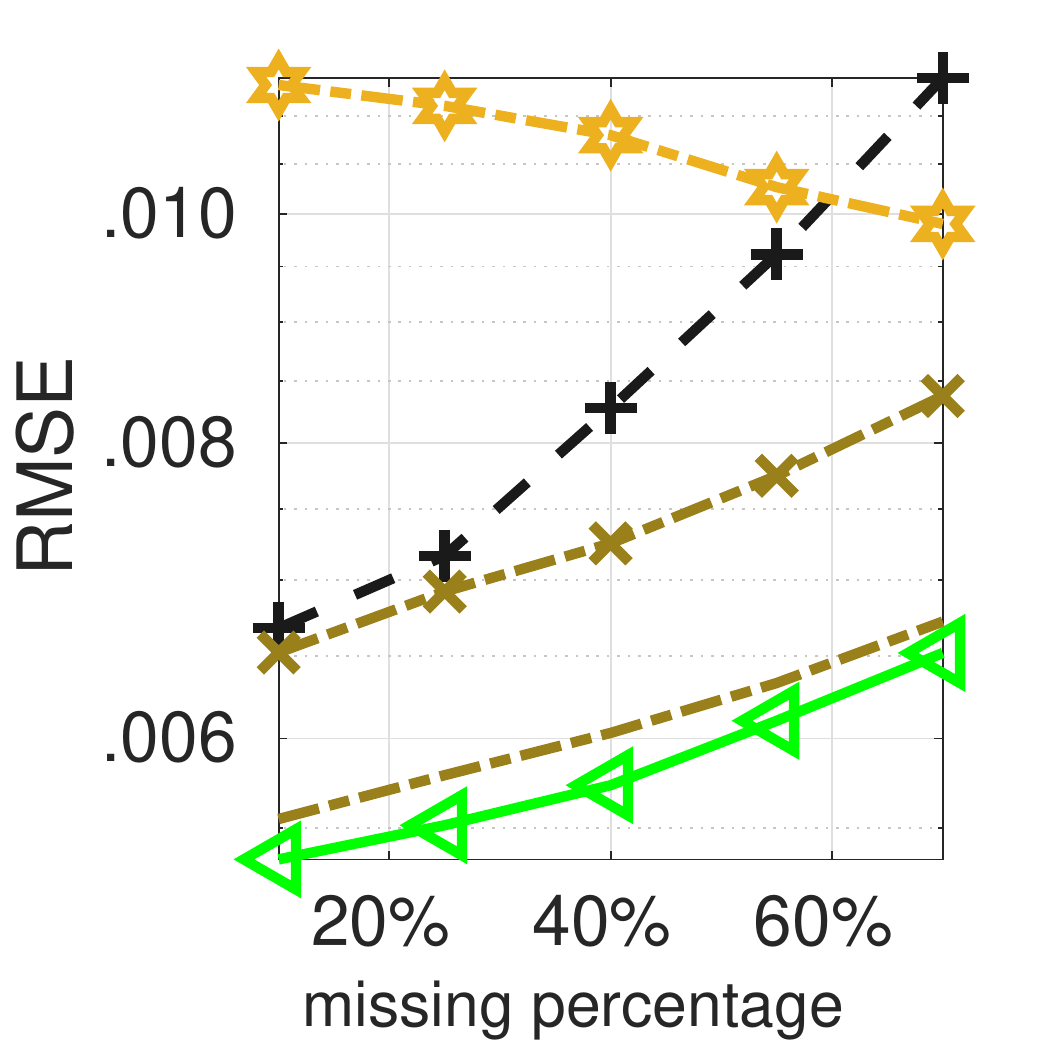}
        \label{fig:traffc_random}
    }\hspace{-0.1in} \vsa\vsa
           \subfigure [  Traffic-interpolate 
           ]
    {
        \includegraphics[width=0.17\textwidth]{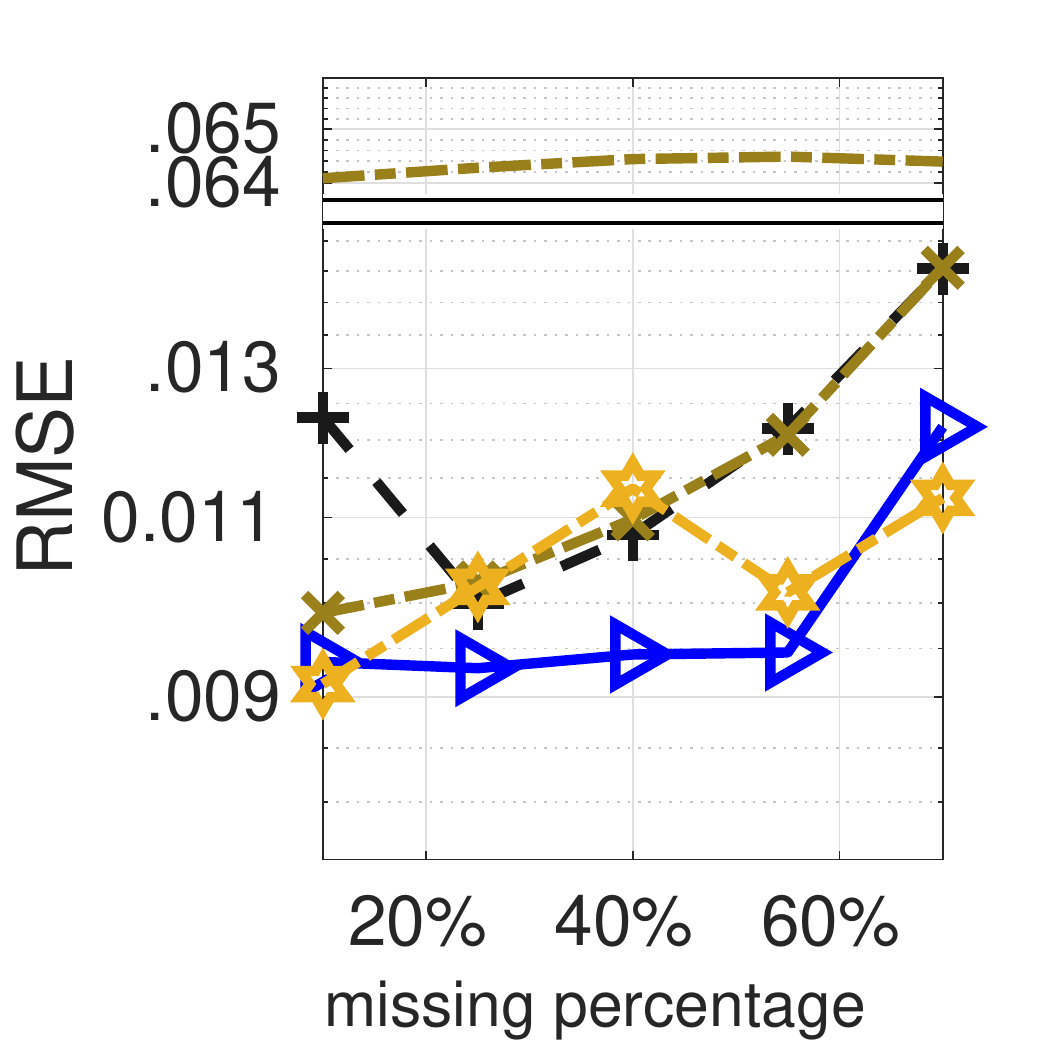}
        \label{fig:traffic_col}
    }\hspace{-0.1in} 
            \subfigure[ Traffic-legend]
      {
        \includegraphics[trim={0 -3.3cm 1 1 },width=0.11\textwidth]{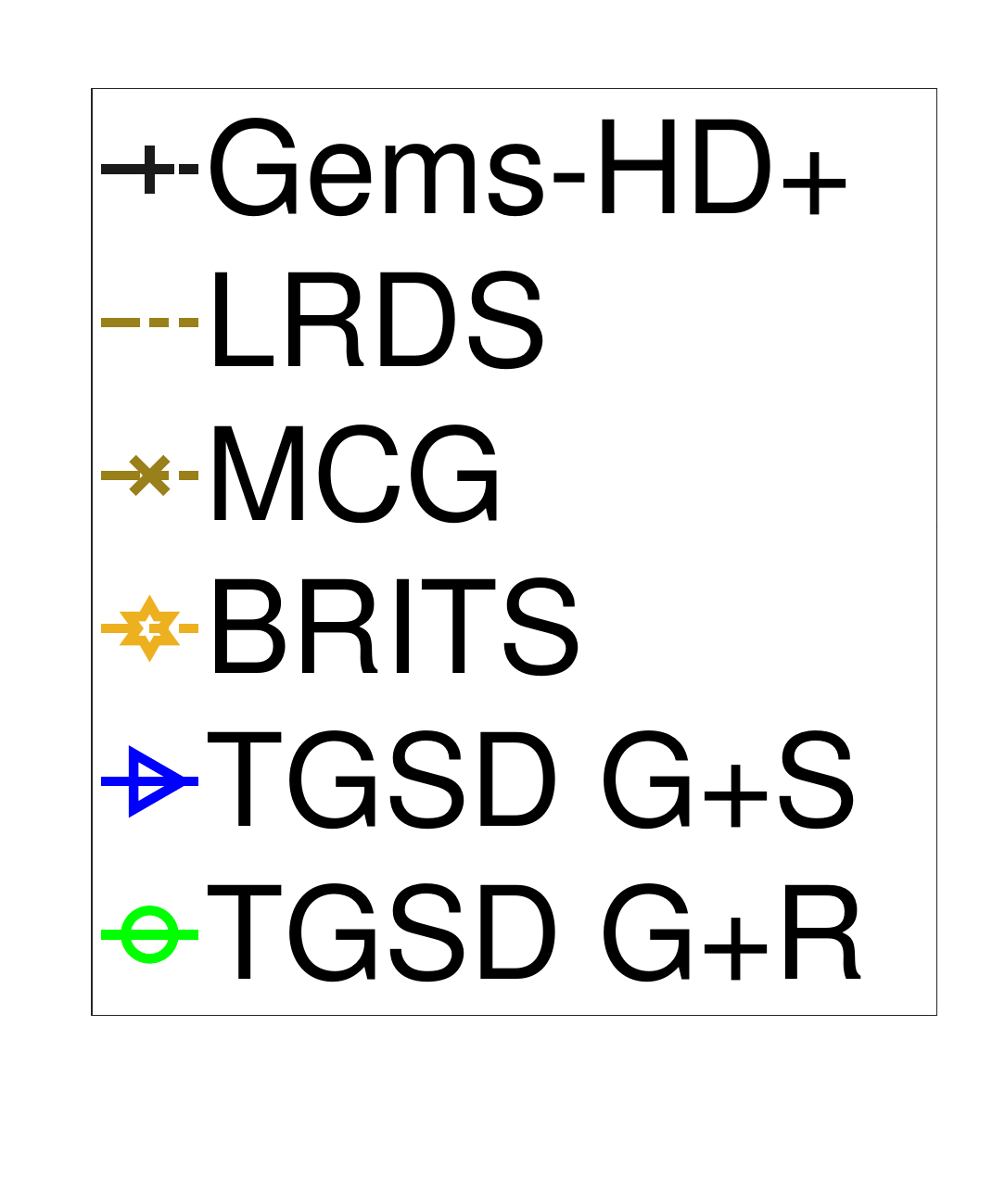}
        \label{fig:traffic_legend}
    }\hspace{-0.1in}

    \caption{\footnotesize Comparison of quality for missing value imputation \subref{fig:syn_random}, \subref{fig:bike_random}, and \subref{fig:traffc_random}; and  for interpolation \subref{fig:syn_col}, \subref{fig:boston_col}, and \subref{fig:traffic_col}}\vsa
    \label{fig:whole_syn_random_missing}\vsa
\end{figure}

       
    
    

\vsb
\subsection{\bf Missing value imputation.}\label{Missing value imputation} Next we evaluate \ourmeth's accuracy in predicting missing values in temporal graph signals. We vary the ratio of missing values by randomly removing observations and quantify the accuracy (as RMSE) of competing techniques to predict them. We perform a dataset-specific parameter search for each method (details in the supplement) and report average accuracy of $5$ sets of removed values for each missing percentage level in Figs.~\ref{fig:syn_random},~\ref{fig:bike_random},~\ref{fig:traffc_random}. 

A version of \ourmeth outperforms all baselines across datasets. Different dictionaries have advantages in specific settings, however. On \textit{Synthetic} data (Fig.~\ref{fig:syn_random}) all dictionaries for \ourmeth outperform baselines, but 
the  GFT + Ramanujan combination dominates since signals have strong periodicity and the Ramanujan encoding is known to be advantageous in such scenarios~\cite{zhang2019PERCeIDs}. MCG and LRDS are the next best methods and follow similar trends. They are both designed for missing value imputation based on low rank and graph smoothness regularizers. Gems-HD+ performs well for small number of missing values but degrades rapidly as the spline-based imputation treats time series independently and does not take advantage of the graph structure which is critical when many values are missing. BRITS's average behavior exhibits an unexpected downward trend, however, the variance of this method is very large with the standard deviation being equivalent 24\% of average RMSE for all runs and especially so for higher missing percentage (27\%). For context \ourmeth G+R had an average standard devation of 1\%. The tens of thousands of parameters in BRIT's model may cause it to overfit noise.
In the \textit{Bike} dataset (Fig.~\ref{fig:bike_random}), the DFT dictionary performs best for \ourmeth and Gems-HD+ is the second best method. This data is aggregated at daily resolution hiding finer temporal patterns and shifting more weight on to the importance of the graph which may explain Gems-HD+ good performance.
LRDS, BRITS and MCG perform significantly worse in this setting. 

In \textit{Traffic} (Fig.~\ref{fig:traffc_random}), similar to synthetic, the Ramanujan dictionary is preferable for \ourmeth and LRDS is the best performing baseline. 
Traffic time series are periodic and smooth along both the graph and time, giving \ourmeth with Ramanujan dictionary an advantage as it can encode both short- and long-term patterns. LRDS and MCG both employ temporal smoothness regularization rendering them the next best methods. \vsa

\vsc
\subsection{Temporal interpolation}\label{Temporal interpolation} We also consider a scenario in which whole temporal slices are missing and quantify the ability of \ourmeth to interpolate in time. 
To this end we remove random slices and present average interpolation RMSE. We again perform a dataset-specific grid search for each methods (detailed in the supplement). 
Similar to value imputation, \ourmeth outperforms all baseline across datasets. The spline temporal dictionary performs on par with the Ramanujan and DFT and better on the Traffic dataset. This is expected as temporal smoothness becomes important when entire snapshots are missing.
In \textit{Synthetic} (Fig.~\ref{fig:syn_col}) all temporal dictionaries perform well, with the Ramanujan having some advantage. The DFT dictionary performance degrades for large number of missing snapshots, while that of Spline and Ramanujan remains stable. BRITS, MCG, and Gems-HD+ also preform well, while the quality of LRDS is the outlier. LRDS smooths out based on values in the same row and column, however, since here whole columns (slices) are missing, its column smoothing is rendered ineffective. 

\ourmeth has a significant advantage over baselines in the \textit{Bike} dataset (Fig.~\ref{fig:boston_col}). The use of spline interpolation as a preprocessing step in Gems-HD+ gives it advantage over other baselines, however, since this preprocessing is graph-agnostic, its performance is inferior compared to \ourmeth. 
While MCG enforces local temporal smoothness, it ignores long-term dependencies such as periodicity, hence its inferior quality on this daily resolution dataset. 
BRITS performs well at both low and high fractions of missing slices in \emph{Traffic} (Fig.~\ref{fig:traffic_col}), even outperforming \ourmeth in those extreme settings by small margins. BRITS' RMSE has a very large standard deviation (12\% of average) across runs indicating potential overfitting to the observed values leading also to a non-smooth average RMSE trend. Among the variants of \ourmeth, the Spline dictionary performs best in this dataset. Gems-HD+ and MCG have similar performance while LRDS performs an order of magnitude worse than all other methods (note that the y axis is discontinuous).

\vsb
\subsection{Clustering node time series} \label{clustering} Since our proposed model computes a representation of nodes in $\Psi Y$, we next seek to evaluate the utility of the latter for node time series clustering on four datasets with ground truth communities: \emph{Synthetic}, \emph{Reality Mining}~\cite{eagle2006reality}, \emph{Reddit-epi}~\cite{zhang2019PERCeIDs} and \emph{Reddit-sp}\cite{zhang2019PERCeIDs}. We compare the performance of \ourmeth, CCTN and PCA in Tbl.~\ref{table:clustering} in terms of clustering accuracy and the running time. All methods uses Kmeans with K equal to the number of clusters to obtain cluster labels.
\ourmeth exhibits better clustering quality in all datasets and scales as well as PCA. 
The periodic dictionaries enable \ourmeth to capture the accurate patterns in time series, resulting in effective features for clustering.
We show two sets of results for CCTN for two values for embedding dimensionality (parameter $d$ in CCTN). The quality value in brackets is for the default $d=3$, while that outside the brackets is for $d$ set to the ground truth number of clusters in each dataset. The performance of CCTN is second best in the \emph{Reality mining} and \emph{Reddit-sp} and it is the slowest alternative among the $3$ alternatives.
Different from \ourmeth, CCTN aims to learn a low-dimensional dictionary to represent all time series. The learned dictionary does not capture interpretable temporal structures such as seasonality, but instead seeks to minimize the reconstruction error in a data-driven manner which may explain its relatively lower performance in the periodic Synthetic dataset. 
The performance of CCTN is also sensitive to the number of embedding dimensions $d$.
PCA is the second best method on the other two datasets and is the fastest among the three. It does not explicitly consider temporal or graph structures in the data and instead seeks to maximally represent the variance within a pre-specified number of components (set to the true number of communities). While PCA is general and fast, its generality becomes a limitation in the presence of rich graph and temporal structures exploited by \ourmeth.






\begin{table}[t]
\setlength\tabcolsep{3 pt}
\centering
\footnotesize
 \begin{tabular}{|c|c|c|c|c|c|c|c|c|c|c|} 
 \hline
  \multicolumn{1}{|c|}{} & \multicolumn{2}{|c|}{TGSD}  &  \multicolumn{2}{|c|}{CCTN~\cite{liu2019coupled}} & \multicolumn{2}{|c|}{PCA} \\
 \hline  
 Dataset & ACC& Time & ACC& Time &ACC& Time\\ \hline
Synthetic & {\bf 85}\% & 0.6s &  14\% (20\%*) & 20s & 57\% &  .02s \\ 
 \hline
   Reality Mining & {\bf 63\%} & 80s & 55\% (50\%*) & 38m & 54\% & 25s \\
 \hline
  Reddit-epi & {\bf 45\%} & 11s &  35\% (32\%*) & 2.2h & 44\% & 4s \\
 \hline
  Reddit-sp & {\bf 38\%} & 37s & 36\%(36\%*) & 4.7h & 35\% & 6s \\
 \hline
\end{tabular}
 \caption{\footnotesize 
 Comparison of TGSD (G+R in Synthetic, G+D otherwise) and baselines on node clustering accuracy (ACC) and running time (Time).
 }\vsa\vsa\vsb
\label{table:clustering}
\end{table}
\vsb
\subsection{Period detection} \label{Period detection}
\begin{figure} [!t]
    \centering
    \subfigure  [Varying noise]
    {
        \includegraphics[width=0.21\textwidth]{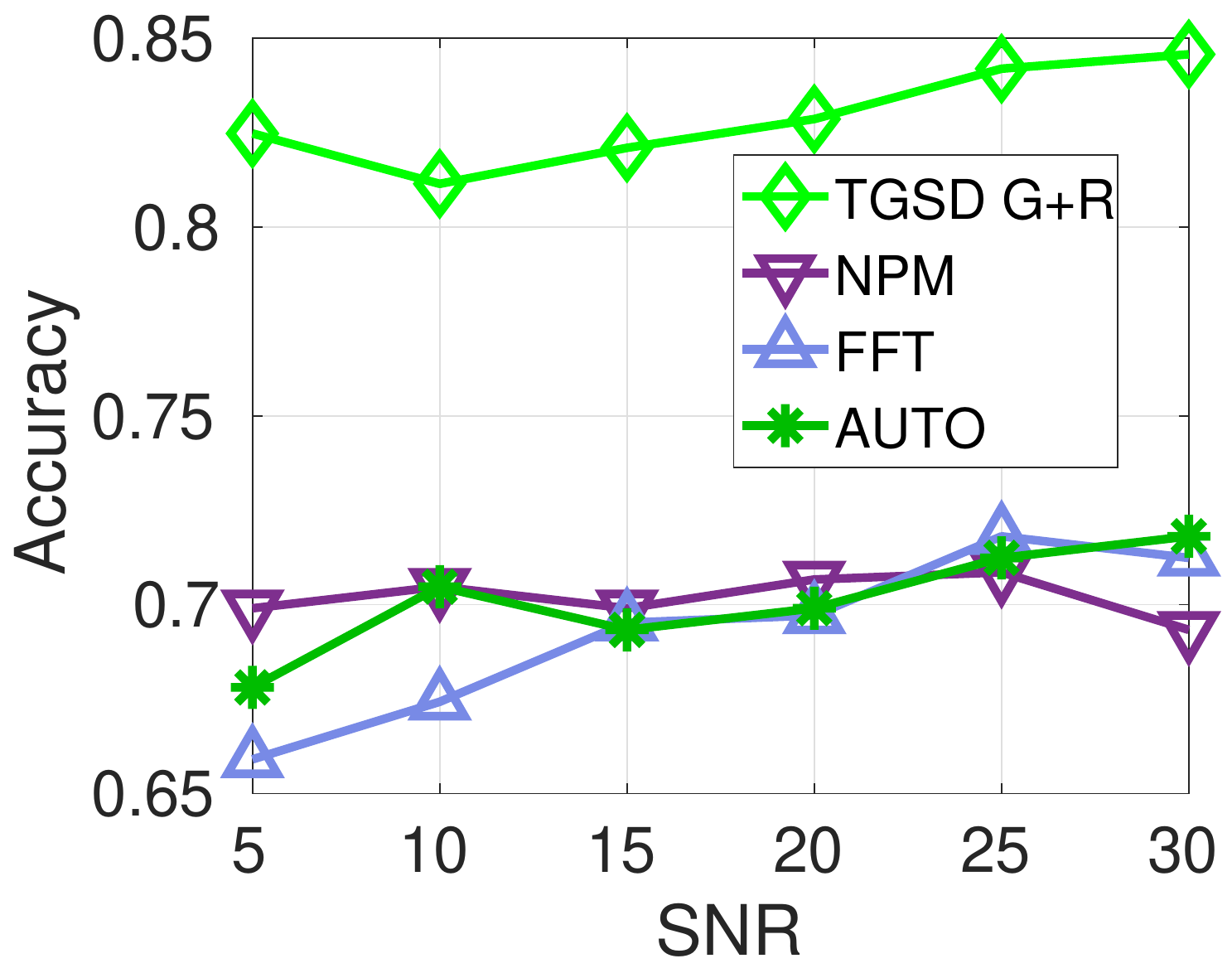}
    }\hspace{-0.1in}
        \subfigure [Varying missing values]
    {
        \includegraphics[width=0.21\textwidth]{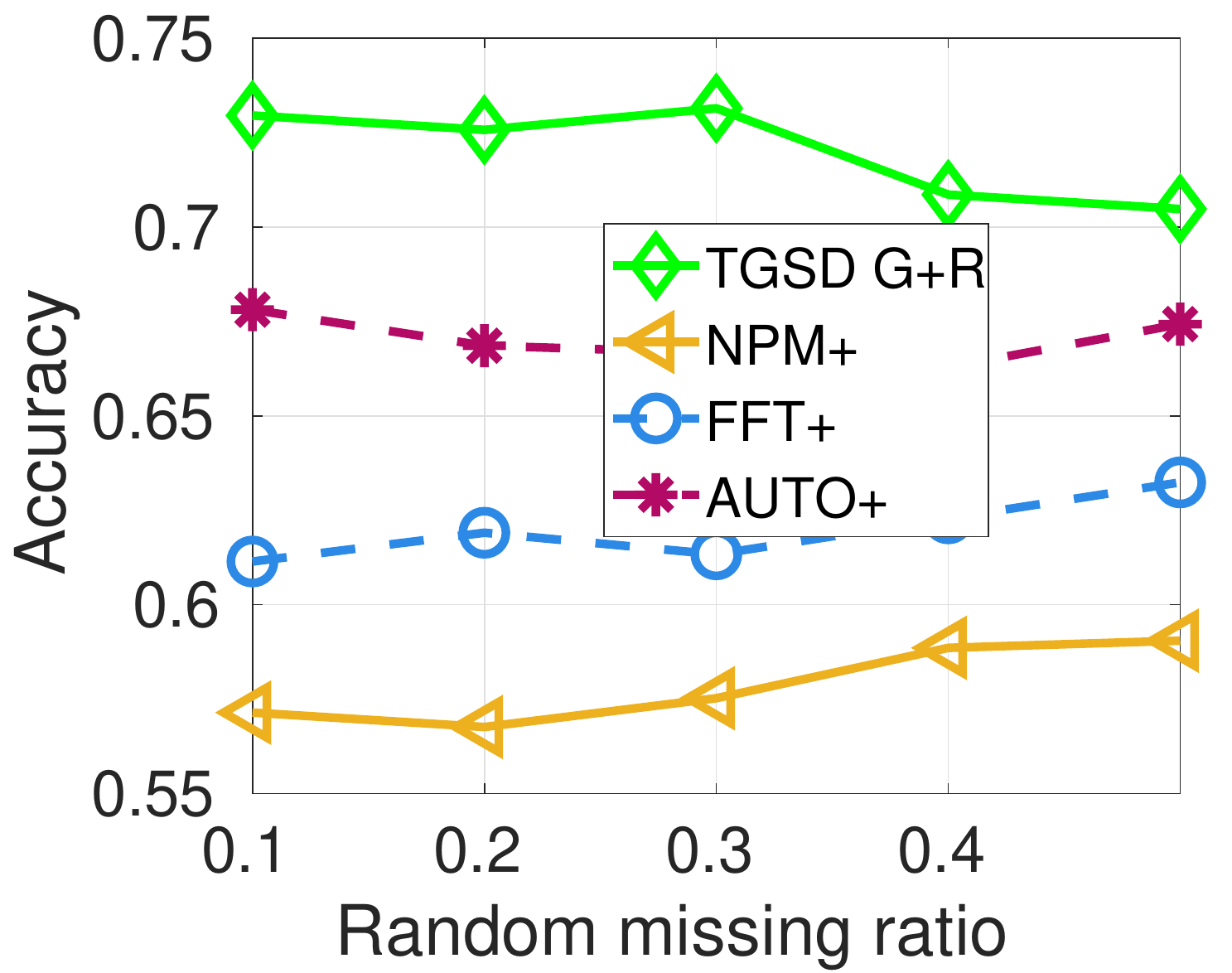}
    }\hspace{-0.1in} \vsa\vsb
    \caption{\footnotesize Period learning in synthetic data. 
    }
    \label{fig:syn_period_learning}\vsa\vsc\vsc
\end{figure}

We also evaluate the performance of \ourmeth for period detection. When using Ramanujan dictionary, 
we can extract a periodic coefficient matrix  $A = \Psi Y W$ and then predict leading periods similar to the approach by Tenneti et Al.~\cite{tennetiTSP2015}. We report the accuracy of the top-$k$ predicted periods in Synthetic data (average of $5$ runs), where $k$ is selected based on the actual number of ground truth (known) periods used to generate the node time series. We perform two experiments by varying (i) the SNR and (2) the percentage of random missing values and report the results in Fig.~\ref{fig:syn_period_learning}. In both experiments, \ourmeth exhibits superior quality: $15\%$ improvement over the best baselines and across SNRs, and up to $10\%$ improvement over the best baseline for varying levels of missing values.
Unlike all baselines which are designed for time series, \ourmeth employs jointly the graph an temporal dictionaries making it less sensitive to noise and more robust in the presence of missing values. For the case of missing values, the graph dictionary employed by \ourmeth allows for better imputation and, in turn, better periodicity estimation. In comparison, baselines employ a two-step approach as they are not designed to handle missing values: they first impute values using spline interpolation in time and then estimate periodicity. Although, baselines employ similar dictionaries for time as those in \ourmeth (NPM employs the Ramanujan dictionary and DFT and AUTO are based on the Fourier transform), their inability to model relationships among individual time series encoded in the graph structure renders them less accurate than \ourmeth.

\begin{figure} [!t]
    \centering
    \subfigure  [Periodic Signal ]
    {
        \includegraphics[width=0.16\textwidth]{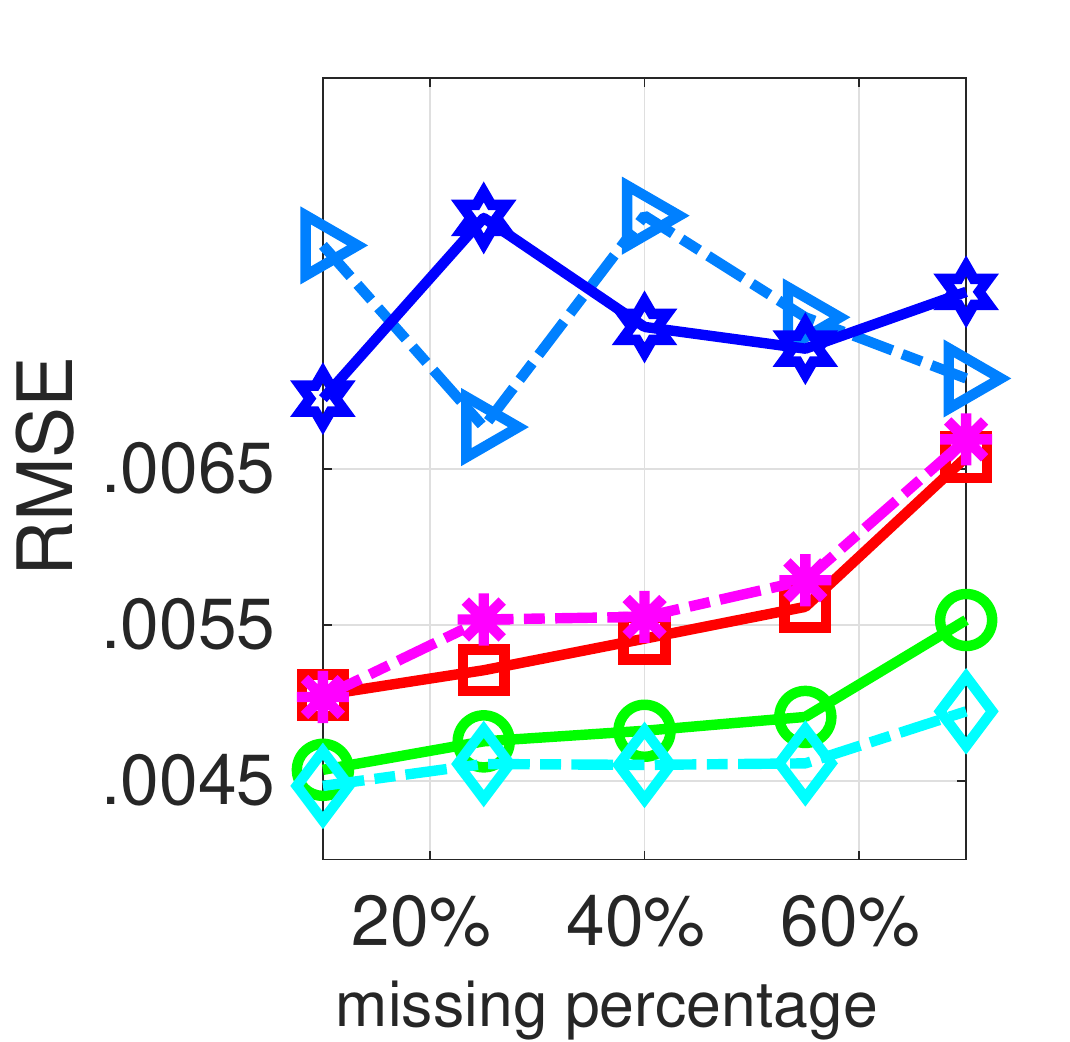}
        \label{fig:period_dic_demo}
    }\hspace{-0.1in}
    \subfigure [ Smooth Signal ]
    {
        \includegraphics[width=0.16\textwidth]{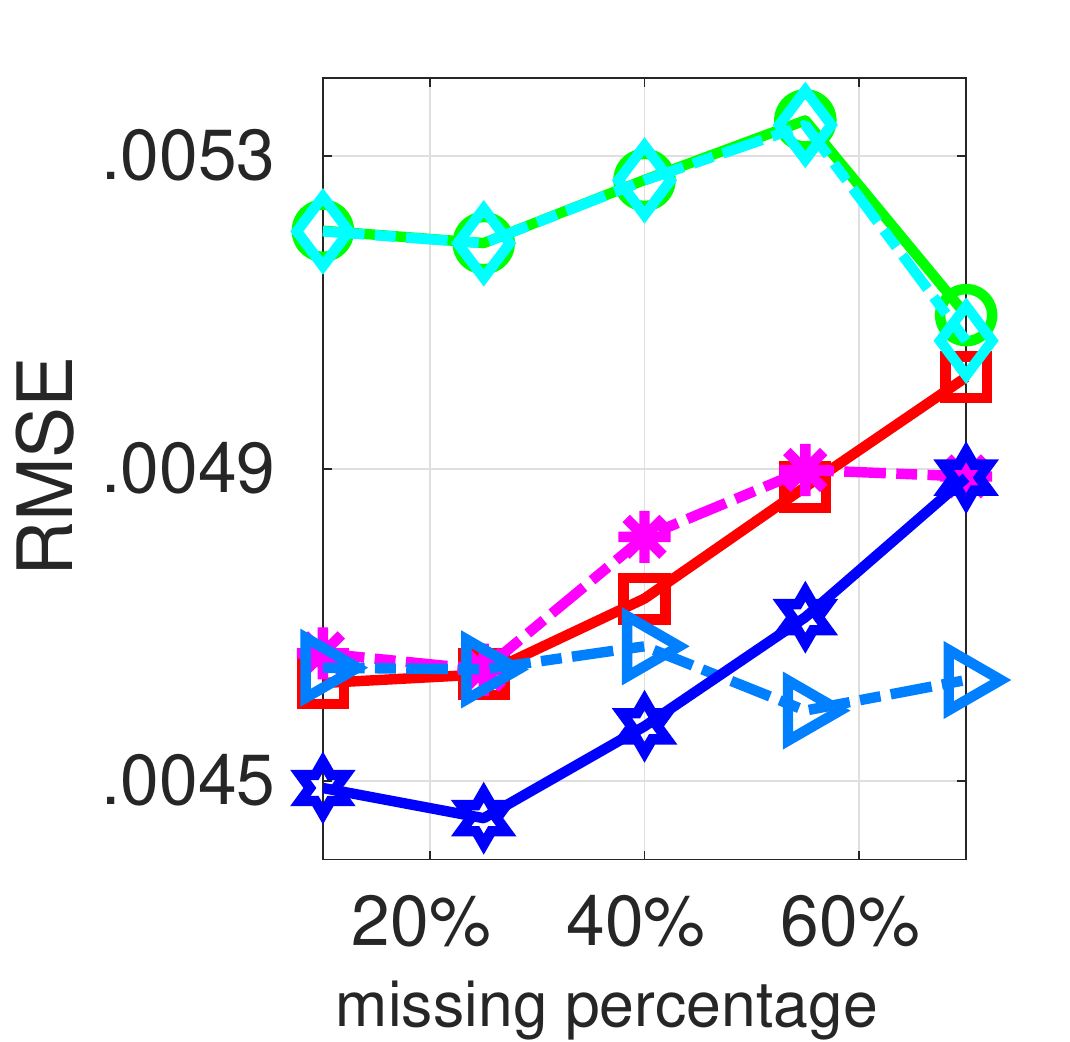}
        \label{fig:smooth_dic_demo}
        
    }
    \hspace{-0.1in}
        \subfigure[Legend]
      {
        \includegraphics[trim={0 -2.7cm 1 1 },width=0.12\textwidth]{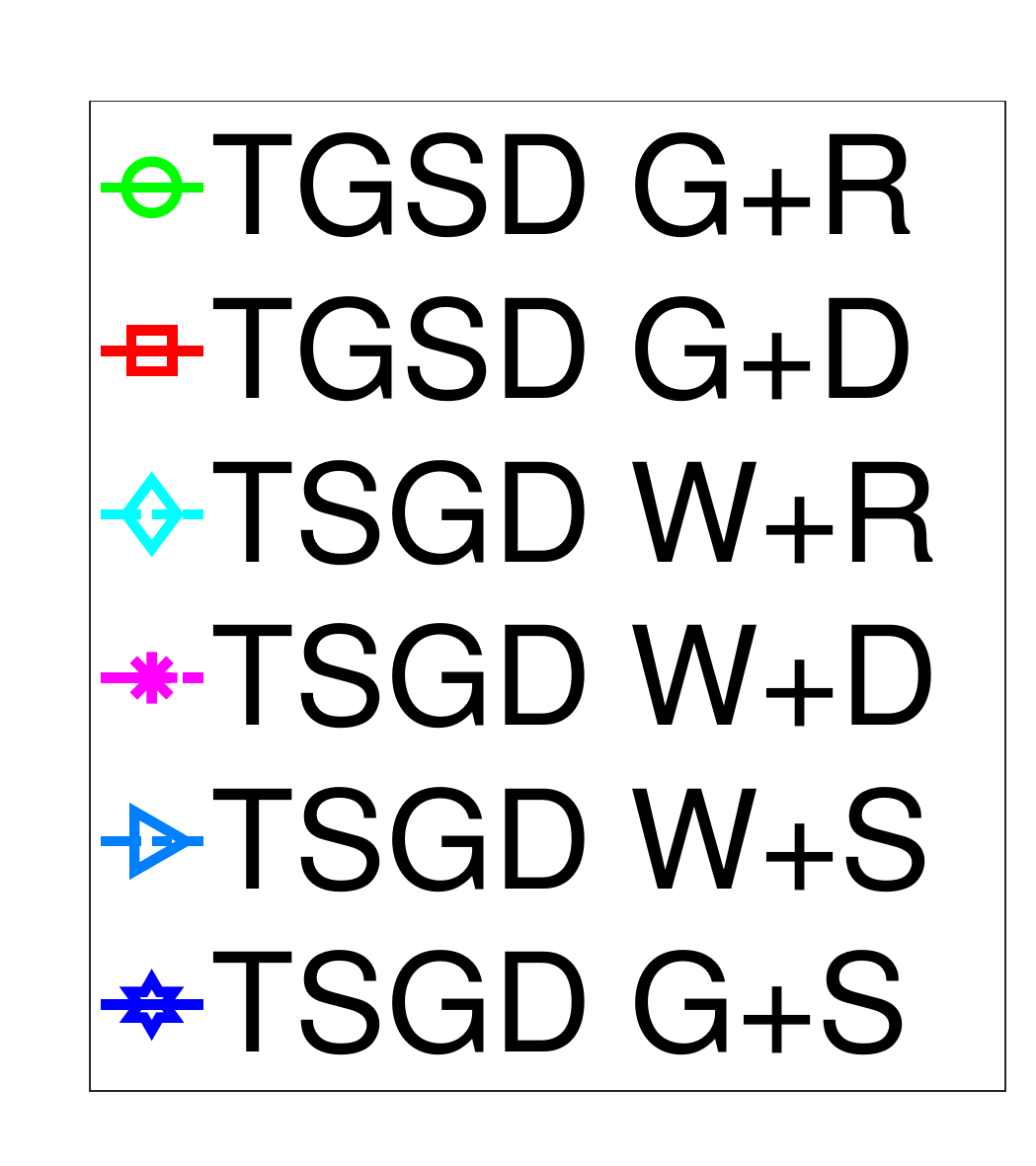}
        \label{fig:dic_demo_legend}
    }\hspace{-0.1in}\vsa\vsc
  \caption{\footnotesize Comparison of dictionary combinations.}
    \label{fig:dic_demo}\vsa\vsa 
\end{figure}

\begin{figure*} [!t]
    \centering
    \footnotesize
\setlength\tabcolsep{2 pt}
         \subfigure[Rank Estimation]
      {
        \includegraphics[width=0.23\textwidth]{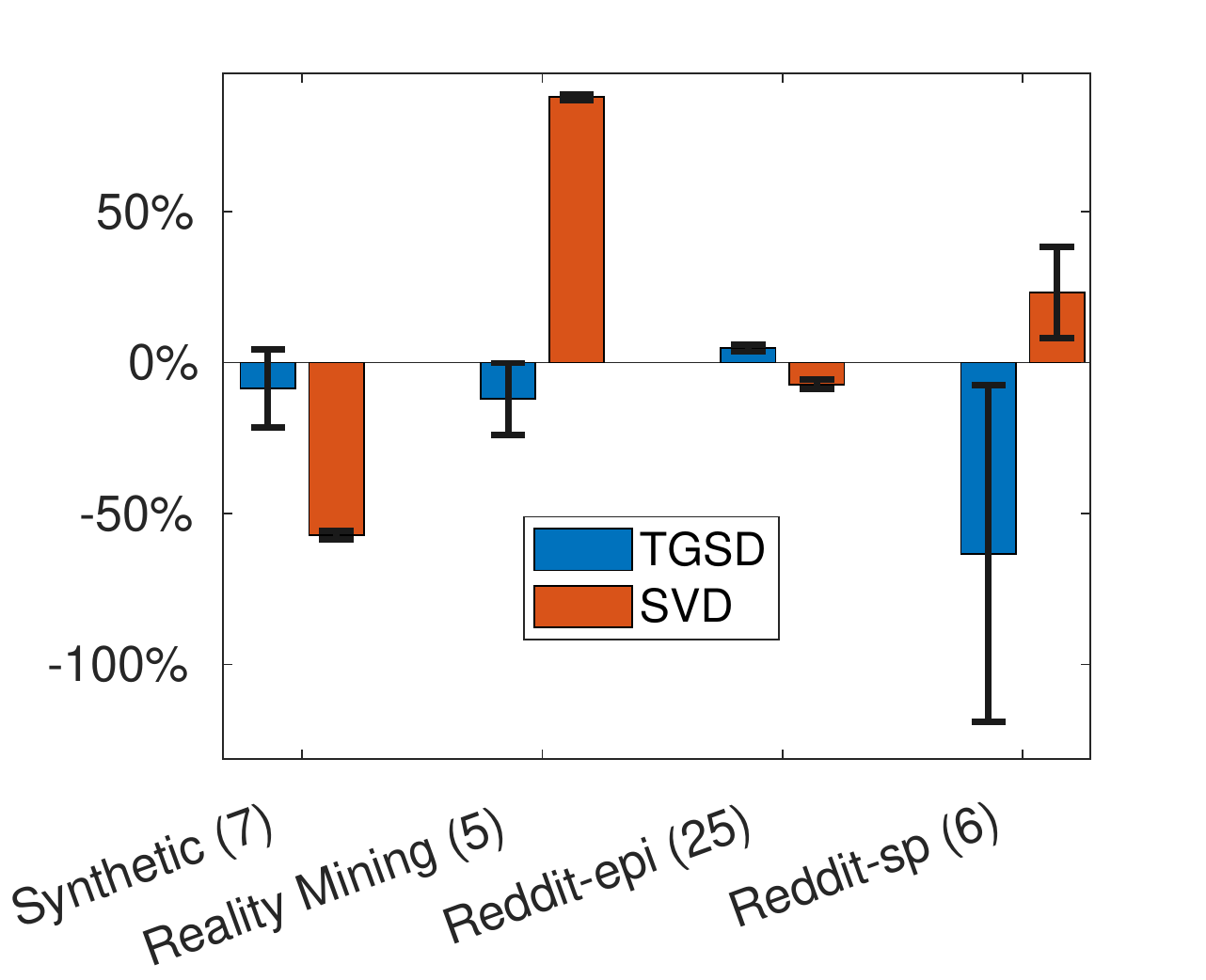}
        \label{fig:rank_est}
    }\hspace{-0.1in}
    \subfigure [Nodes]
    {
        \includegraphics[width=0.16\textwidth]{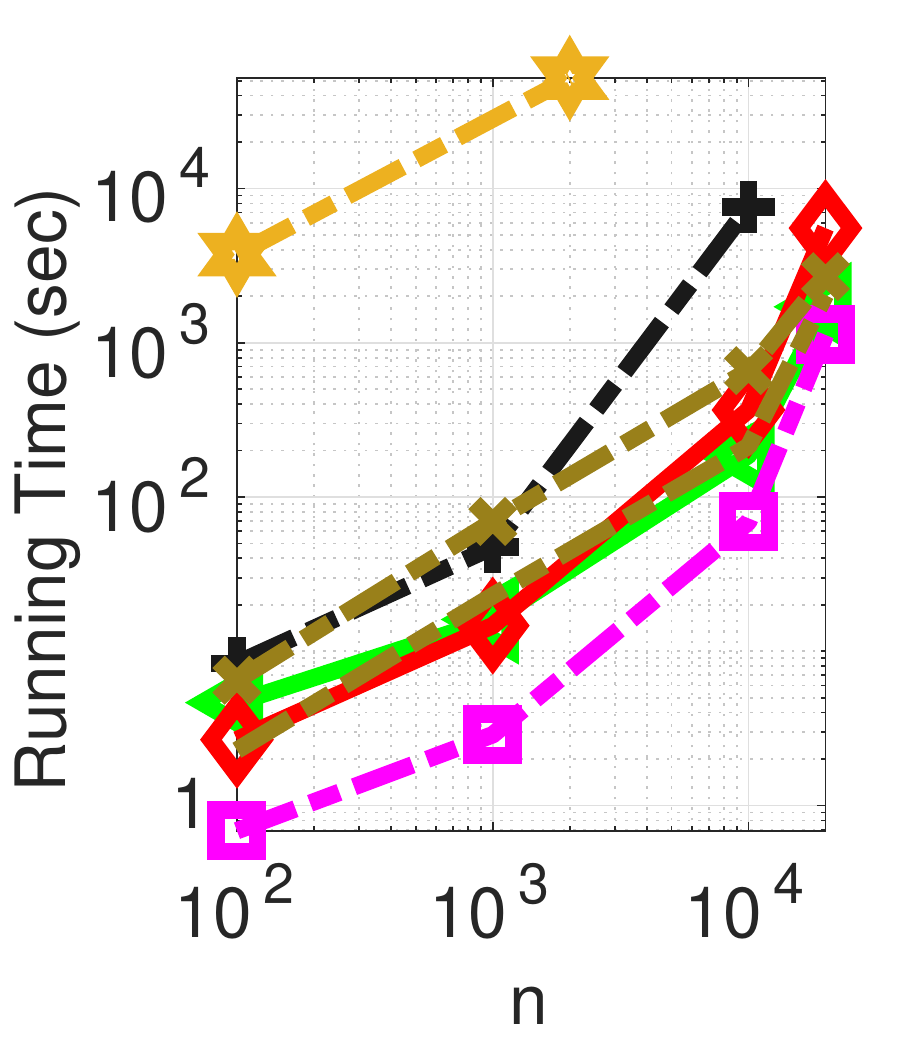}
        \label{fig:run_time_varying_nodes}
    }\hspace{-0.1in}
    \subfigure[Timestamps]
      {
        \includegraphics[width=0.16\textwidth]{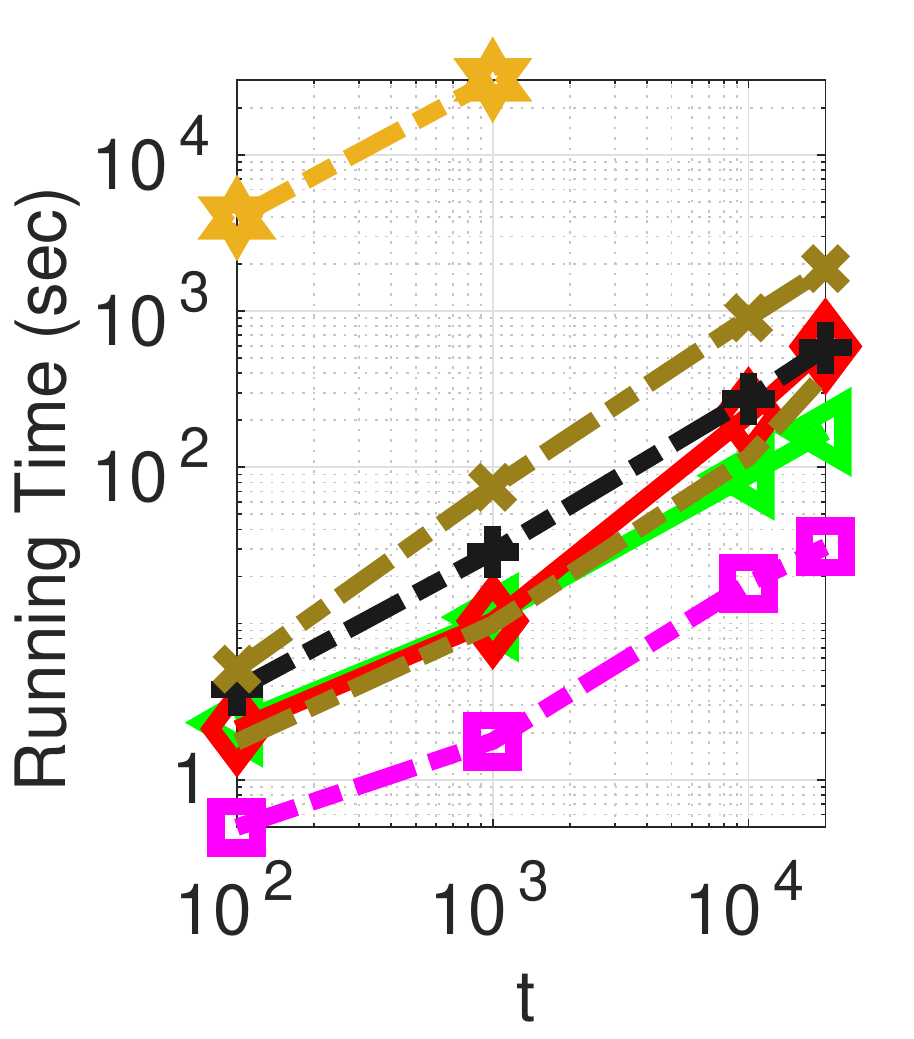}
        \label{fig:syn_running_time_aurora}
    }\hspace{-0.1in}
        \subfigure[Legend Scalability]
      {
        \includegraphics[trim={0 -7.5cm 1 1 },width=0.13\textwidth]{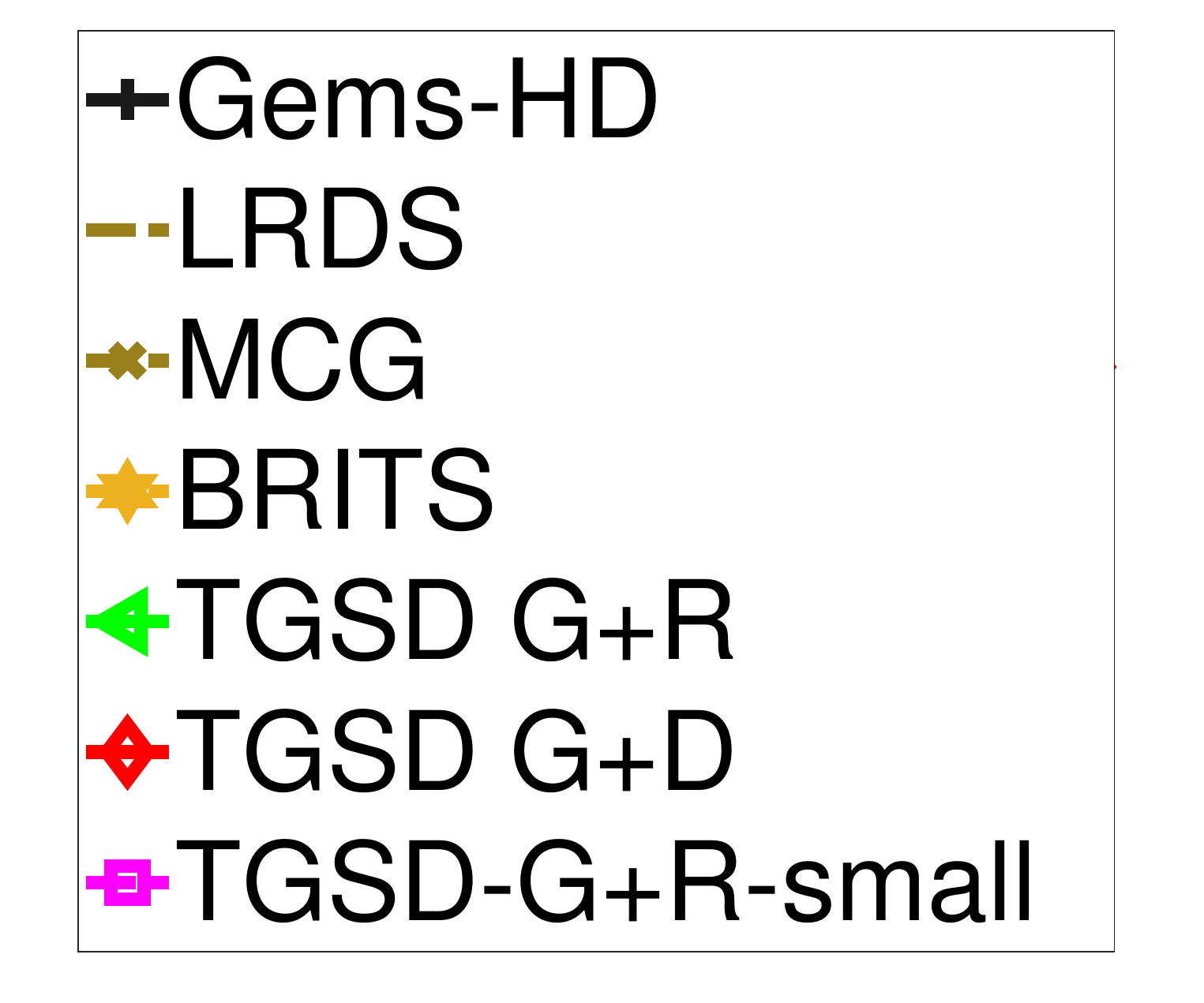}
        \label{fig:syn_running_time_legend}
    }\hspace{-0.1in}
     \subfigure[Effect of partial dictionary]
      {
        \includegraphics[width=0.24\textwidth]{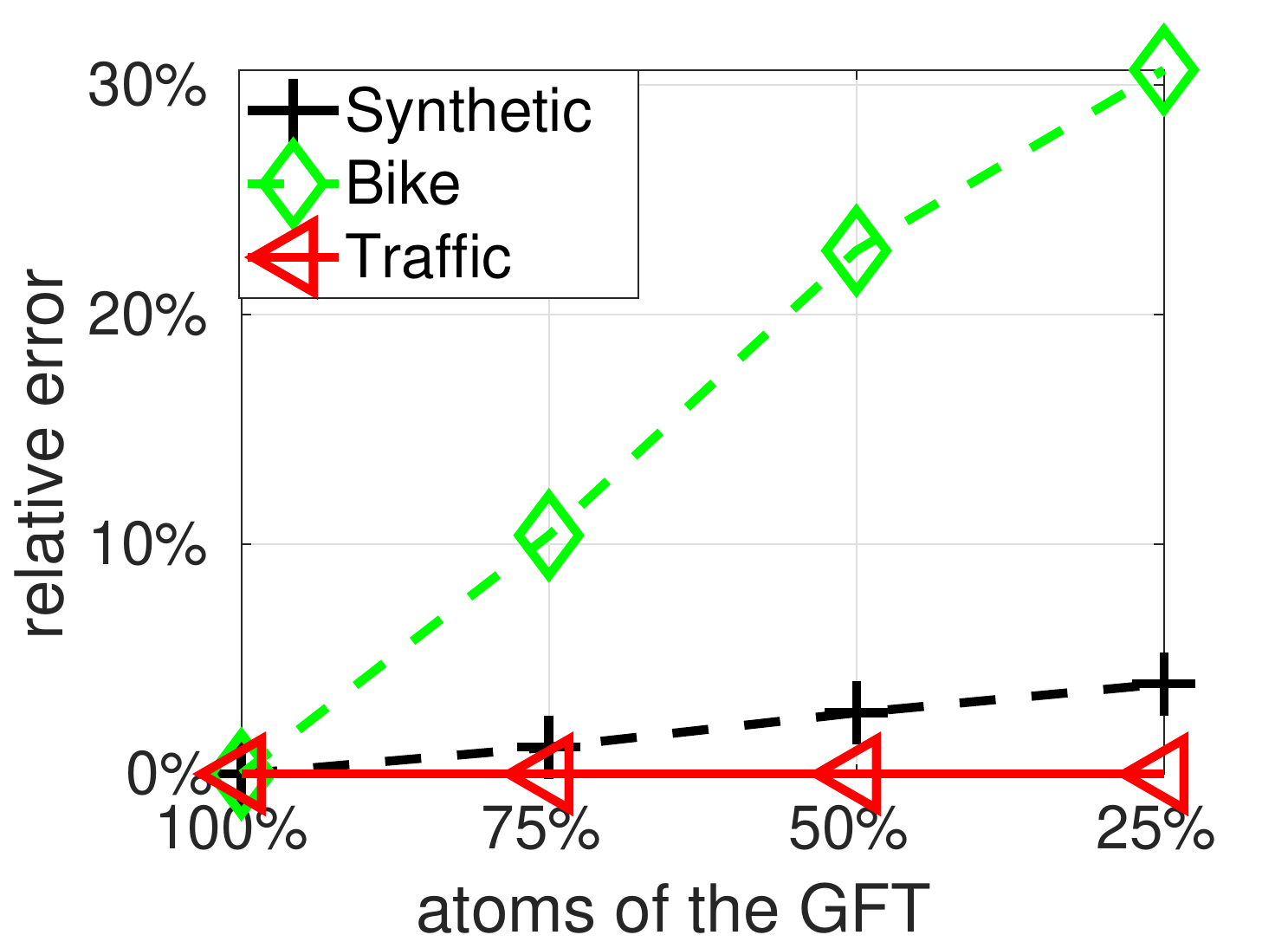}
        \label{fig:atom_result}
    }\hspace{-0.1in}
    \vsa\vsa
    \caption{\footnotesize Rank detection relative error and variance (\ourmeth G+R for Synthetic \ourmeth G+D for all others), scalability comparison as a function of (b) $n$ and (c) $t$; and (d) quality of \ourmeth value imputation for partial GFT dictionary.}
    \label{fig:time_sensitive}\vsa
\end{figure*}

\vsa
\subsection{Dictionary comparison} 
We compare the dictionary combinations for the missing value imputation task on synthetic data in Fig.~\ref{fig:dic_demo}. In Fig.~\ref{fig:period_dic_demo} we generate the node signals in the same manner as in the \textit{Synthetic} datasets employed in previous experiments, namely we allow nodes within clusters to share periodic trends. As a result, the Ramanujan dictionary coupled with both GFT and Wavelets in the graph domain performs the best. In contrast, in Fig.~\ref{fig:smooth_dic_demo} we generate node signals using random (non-periodic) and smooth trends. As expected, for such locally smooth but non-periodic data, the Spline dictionary works best for encoding the temporal domain. The DFT dictionary provides a middle ground in terms of quality in both cases. 

\subsection{Estimation of the number of components $k$} \label{estimat rank}

Another important parameter for \ourmeth is the number of components $k$ for the decomposition. Setting this parameters is a similar problem to determining an optimal rank for other decomposition approaches (e.g., PCA, SVD, NMF), for which a cross-validation was shown to perform best among alternatives based on statistical tests and heuristics~\cite{cv_rank_est}. In the element-wise k-fold (ekf) cross-validation folds are values are randomly created and removed from a matrix, then imputed and predicted by a reconstruction method with different number of components $k$~\cite{cv_rank_est}. The $k$ leading in the lowest average SSE is predicted as the optimal rank. We extend this rank estimation protocol to \ourmeth and compare the quality of estimating $k$ using \ourmeth and SVD~\cite{cv_rank_est} employing spline-imputation on \emph{Synthetic}, \emph{Reality Mining}~\cite{eagle2006reality}, \emph{Reddit-epi, Reddit-sp}~\cite{zhang2019PERCeIDs}. 

We compare the quality of the two estimates w.r.t. number of ground truth communities employing 5-fold cross validation and present results in~\ref{fig:rank_est}. \ourmeth outperforms SVD in all but the \emph{Reddit-sp} datasets. \ourmeth is able to handle missing values directly meaning that is not affected by the quality of a reprocessing imputation scheme used in SVD. It is also able to utilize the graph and temporal dictionaries to further guide its missing value imputation. A similar strategy can be adopted for \ourmeth's other parameters $\lambda_1$ and $\lambda_2$. 

\eat{
We have two categories of temporal dictionary, periodic and smooth. 
The two periodic dictionaries are of course the DFT and Ramanujan. These basis differ in a few key ways.
The first is that Ramanujan dictionary excels on data with strong periodic trends. This is most clear in the synthetic where the Ramanujan almost always outperformed DFT. In the real world we found the Bike data set was less periodic so DFT always performed better but  the CCT data set was more mixed with period and other trends thus DFT performed better for some task and DFT for others. 
The second difference is that Ramanujan requires a max period as input where DFT is parameter free. In the synthetic we were able to set the max period appropriately because we knew what the actual max period was (we knew it was 7 and set the Ramanujan to 10).
The third and related point is that dictionaries have different computation complexities. The DFT matrix is orthogonal and thus we can use the easier optimization method laid out in \todo{Cite formula}. The second is that for the Ramanujan the number of free parameters that need to be solved for in $W$ is variable depending on the setting of the max period. This is because the rows of a Ramanujan grows in a non linear way as a function of the max period. For example at a max period of 10 a Ramanujan is $30\times t$ but at a max period of 50 the Ramanujan is $774\times t$. \todo{so Which is better in complexity?}.
The other dictionary we use is a Spline dictionary. Spline Dictionaries will enforce general smoothness on our reconstruction. We see that Spline imputation works well on column imputation. 
We suggest to use Ramanujan when a signal is known to have a strong periodic signal with a known max period, DFT when there is some latent periodicity, and spline for just general temporal smoothness. 
}

\vsb
\subsection{Scalability and partial graph dictionaries.}
We next evaluate the scalability of competing techniques with the number of nodes and time steps. We add a partial dictionary variant of TGSD, \ourmeth-G+R-small, in the scalability comparison to quantify the scalability improvement it may offer. Namely, \ourmeth-G+R-small employs $10\%$ of the lowest-frequency GFT atoms, i.e. the Laplacian eigenvectors corresponding to the smallest eigenvalues.

\ourmeth-G+R-small scales best among alternatives, while \ourmeth with the full GFT and Ramanujan dictionaries are the second fastest approach (Fig.~\ref{fig:time_sensitive}). \ourmeth with GFT and DFT is the slowest among \ourmeth's variants due to the relatively large dictionaries (the full DFT is $\Phi\in\mathcal{R}^{(t\times t)}$ and the full GFT: $\Psi\in\mathcal{R}^{(n\times n)}$). Among the baselines, \textbf{MCG} scales significantly worse than \textbf{LRDS} with both $t$ and $n$ since the time and graph regularizers employed in the latter retain convexity of the overall objective and allow for a more efficient solver. In contrast, \textbf{MCG} employs a nuclear norm to enforce a low rank, resulting in a $O(nt^2 + t^3)$ worst-case complexity, where $t$ is the number of columns. 
\textbf{Gems-HD}'s running time grows quickly with the number of nodes as it employs the graph-Haar wavelets for basis, resulting in a cost of $O(n^2)$ per atom (atom encodings are fit one at a time). \textbf{BRITS}, which is based on deep learning, scales orders of magnitude worse than all other methods as it needs to learn many more parameters requiring hundreds of epochs. It is important to note that while all other methods are executed on conventional CPU architecture, the running time results we report for \textbf{BRITS} are for execution on a dedicated GPU server with state-of-the-art NVIDIA Tesla V100 GPU card. 

While a partial graph dictionary offers scalability improvement (\ourmeth-G+R-small in Figs.~\ref{fig:run_time_varying_nodes},~\ref{fig:syn_running_time_aurora}), it is natural to question if these savings come at the expense of quality. To investigate this trade-off, we compare \ourmeth's quality on  missing value imputation with decreasing subset of the GFT dictionary atoms. 

The GFT has ordered columns of frequencies with the first column corresponding to the lowest "frequency" and following columns corresponding to finer partitions, i.e. higher frequencies. We leverage this information by only employing a fixed percentage of the lowest frequency atoms for encoding. We report results for \ourmeth in Fig.~\ref{fig:atom_result} for random value imputation at $25\%$ missing values in the \emph{Synthetic}, \emph{Bike} and \emph{Traffic} datasets. The only data set that is significantly impacted by using a reduced GFT is \textit{Bike} while the quality on Synthetic and Traffic is practically unaffected when employing as few as $25\%$ of the atoms. The \textit{Bike} has the smallest graph making every column in the GFT important. In contrast the \textit{Traffic} dataset has by far the largest graph and signals on it can be encoded using very few atoms due to strong local similarity of node behavior (neighboring nodes are consecutive sensors on the same highway and they observe similar speed).

The promising results on scalability and retained quality for partial dictionaries suggest that there exists a significant opportunity for further time savings for \ourmeth by carefully selecting partial dictionaries for both time and the graph structure. 
Beyond partial dictionaries and how to sub-select them, automatic dictionary type selection and optimal selection of the number of components $k$ also promise further reduction in the running time and quality improvement for our method. Such considerations is beyond the scope of this paper, though a promising direction for future investigation.

\section{Conclusion}
In this paper we proposed a general framework for dictionary-based decomposition of temporal graph signals, called \ourmeth. Our algorithm employs an ADMM optimization procedure and can take advantage of multiple existing dictionaries to jointly encode the time and graph extents of temporal graph signals. We performed an exhaustive evaluation of \ourmeth on five application tasks and demonstrated its effectiveness in both synthetic and real-world datasets. 
\ourmeth dominated baselines across tasks. In particular, TGSD achieved $28\%$ reduction in RMSE compared to baselines for temporal interpolation of graph signals when as many as $75\%$ of the observed snapshots were missing. At the same time, \ourmeth scaled best with the size of the input with the fastest variation processing 3.5 million data points in under 20 seconds   while producing the most parsimonious and accurate decomposition models. 

\begin{acks}
This research is funded by an academic grant from the National Geospatial-Intelligence Agency (Award No. \# HM0476-20-1-0011, Project Title: Optimizing the Temporal Resolution in Dynamic Graph Mining). Approved for public release, 21-302. The work is also partially supported by the NSF Smart and Connected Communities (SC\&C) grant CMMI-1831547.
\end{acks}

{\footnotesize 
\bibliographystyle{abbrv}
\bibliography{references,pie_reference,aurora_reference,dsl_ref}
}

\appendix

\clearpage\newpage


\begin{table}[b]
\footnotesize
\setlength\tabcolsep{3 pt}
\centering
 \begin{tabular}{|l|c|c|c|} 
  
    \multicolumn{2}{c}{{\bf BRITS}}  \\
  \hline
\emph{Data/Task} & \emph{Syn. and Bike / imputation and interpolation}  \\
\hline
batch size & \{54,108,216\}  \\
\hline
epoch & \{500,1000,1500\} \\
\hline
hidden layer size & \{34,64,128\}\\
\hline \hline
\emph{Data/Task} &  \emph{Traffic / imputation} \\
  \hline
 batch size  & \{54\}  \\
\hline
epoch & \{500,1000,1500\} \\
\hline
hidden layer size & \{34,64,128\} \\
\hline \hline
\emph{Data/Task}  & \emph{Traffic / interpolation}\\
\hline
batch size & \{54\}\\
\hline
epoch &  \{500,1000\} \\
\hline
hidden layer size & \{34,64,128\}\\
  \hline

\multicolumn{2}{c}{{\bf LRDS}}  \\
 \hline
 \emph{Data/Task} & \emph{All / decomposition, imputation, interpolation}   \\
 \hline
 Nuclear norm weight & \{.1,.5,1\} \\
 \hline
Smoothness weight & \{.1,.5,1\} \\
  \hline
Convergence weight & \{.1,.5,1\}\\
   \hline \hline
 \emph{Data/Task} & \emph{All / only decomposition}   \\
 \hline
 ``zero'' threshold & \{.0001,.001,.01,.1\} \\
\hline
    \multicolumn{2}{c}{ {\bf MCG}}  \\
  \hline
 \emph{Data/Task} & \emph{All / decomposition, imputation, interpolation}   \\
 \hline
 Nuclear norm weight & \{.1,.5,1\} \\
 \hline
Column graph weight & \{.01,.1,1 \} \\
  \hline
Row graph weight & \{.01,.1,1\}\\
   \hline \hline
   \emph{Data/Task} & \emph{All / only decomposition}   \\
 \hline
    ``zero'' threshold & \{.0001,.001,.01,.1\} \\
\hline
         \multicolumn{2}{c}{{\bf GEMS-HD}} \\
  \hline
 \emph{Data/Task} & \emph{Synthetic /  decomposition}  \\
 \hline
 Target atom sparsity & [1:10:100] \\
 \hline
Target signal sparsity & [1:10:100] \\
  \hline
Dictionary size & [1:10:100] \\
   \hline \hline
    \emph{Data/Task} & \emph{Bike /  decomposition}  \\
 \hline
 Target atom sparsity & [1:10:100],[100:20:200] \\
 \hline
Target signal sparsity & [1:10:100],[100:20:200] \\
  \hline
Dictionary size & [1:10:100],[100:20:200] \\
   \hline \hline
  \emph{Data/Task} & \emph{Traffic /  decomposition}  \\
 \hline
 Target atom sparsity & [1:10:100],[100:20:200],[100:20:220] \\
 \hline
Target signal sparsity & [1:10:100],[100:20:200],[100:20:220] \\
  \hline
Dictionary size &[1:10:100],[100:20:200],[100:20:220] \\
   \hline
        \hline
    \multicolumn{2}{c}{{\bf GEMS-HD+}} \\
  \hline
 \emph{Data/Task} & \emph{All / imputation, interpolation}   \\
 \hline
 Target atom sparsity & \{5,10,15,20,25,30\} \\
 \hline
Target signal sparsity & \{5,10,15,20,25,30\} \\
  \hline
Dictionary size & \{5,10,15,20,25,30\} \\
   \hline
    
    \multicolumn{2}{c}{{\bf TGSD}} \\
 \hline
 \emph{Data/Dictionaries/Task} & \emph{all/ all/ decomposition} \\
 \hline
  $\lambda_1$  & \{.001,.01,.1,1\} \\
 \hline
$\lambda_2$ & \{.001,.01,.1,1\}, \\
   \hline
   $k$ & [1:10],[10:10:100] \\
  \hline
 $\%$ of $\Psi$ atoms used & [10\%:10\%:100\%]  \\
  \hline \hline
 \emph{Data/Dictionaries/Task} & \emph{all/ all/ imputation, interpolation} \\
 \hline
  $\lambda_1$  & \{.01,.1,1,10\} \\
 \hline
$\lambda_2$ & \{.01,.1,1,10\}, \\
  \hline
 $\lambda_3$ &  \{.1,1,10\} \\
   \hline
   $k$ & \{5,10,15,20,25,30\} \\
   \hline
\end{tabular}
\caption{\footnotesize Parameter search space for each data set and competing methods in decomposition, imputation and interpolation experiments. Curly braces (e.g. $\{1,2,3\}$) indicate a set of values we tested and brackets (e.g. $[1:2:9]$) indicate that we iterate in an interval from the first to the third values by a step specified in the middle value ( $[1:2:9]$ represents the values $\{1,3,5,7,9\}$)}
\label{table:grid_all}\vsa\vsa\vsa\vsa
\end{table}

\section*{Supplement}

In what follows we discuss how parameters were set in each experiment in the following order: imputation and interpolation, graph signal decomposition, clustering, then period detection. Our goal in providing these experimental details is to facilitate reproduciblity.   
Tbl.\ref{table:grid_all} contains the search grid for all methods and their parameters used in graph signal decomposition (Sec \ref{text:graph signal decomposition}), imputation (Sec \ref{Missing value imputation}) and interpolation (Sec \ref{Temporal interpolation}) experiments. Tbl.~\ref{table:grid_resut} shows the estimated optimal parameters for baselines while Tbl.~\ref{table:tgsd_parm_resut} shows the parameters we used in \ourmeth for all experiments.




\noindent \textbf{$\bullet$ Imputation (Sec \ref{Missing value imputation}) and Interpolation (Sec \ref{Temporal interpolation}). } 
For all method used in imputation and interpolation experiments we preform a task- and data-specific grid search at $25\%$ missing values and use the best parameters for all missing value levels. 
The specific values searched are shown in Tbl. \ref{table:grid_all}.
For BRITS's and LRDS's parameter search spaces we check values both higher and lower than the default parameters set by the authors in the codes they kindly have made publicly available. Due to scalability issues the searched ranges for BRITS are coarser in larger datasets (Tbl. \ref{table:grid_all}) as finer searches did not complete within weeks on state-of-the-art GPU servers. For MCG we explore parameters similar to those for LRDS as they are both low rank methods with smoothing. The authors of GEMS-HD+ list optimal parameters in their publication and we tests alternatives ``around'' these prescribed values. We report the best found parameters for each experiment in Tbl.~\ref{table:grid_resut}. 

\noindent \textbf{$\bullet$ Data decomposition (Sec \ref{text:graph signal decomposition})}.
The parameter settings used in graph signal decomposition can also be found in Tbl.~\ref{table:grid_all}. For MCG and LRDS we search the same parameter spaces and add a "zero" threshold parameter which allows us to control the size of the models. 
For GEMS-HD we expand the search space to obtain a clearer trend for the number of nonzero coefficients necessary for a wide range of RMSE. We test a larger range of values for larger datasets as they require a larger model sizes (nonzero coefficients). We do not include an optimal value table for this search space as the optimal parameters will vary with RMSE and NNZ.

\begin{table}[t!]
\footnotesize
\setlength\tabcolsep{3 pt}
\centering
 \begin{tabular}{|l|c|c|c|c|c|c|c|c|} 
  \hline
  Data-Task & Variation & $\lambda_1$ & $\lambda_2$  & $\lambda_3$  & $k$ \\
  \hline
 \multirow{3}{*}{Synthetic-imputation} & TGSD W+D & 0.01 & 1 & 1 & 5
 \\
 \cline{2-6}
   & TGSD G+D & 0.01 & 0.1 & 1 & 5
 \\
 \cline{2-6}
   & TGSD G+R & 0.1   & 0.1 & 1 & 5
 \\
 \hline
\multirow{3}{*}{Synthetic-interpolation} & TGSD G+S & 1 & 1 & 1 & 5
    \\
 \cline{2-6}
 & TGSD G+D & 1 & 1 & 1 & 5 
    \\
 \cline{2-6}
 & TGSD G+R & 0.01 & 1 & 1 & 30 
    \\
     \hline
 Synthetic-clustering & TGSD G+R & 10 & .01 & NA & 7 
    \\
    \hline
 Synthetic-period detection & TGSD G+R & .1 & .1 & 10 & 1 
\\
 \hline
  Synthetic-k estimation & TGSD G+R & .01 & .001 & 10 & Vary
\\
 \hline
 \multirow{2}{*}{Bike-impute} & TGSD W+D & 0.01 & 0.1 & 10 &25 
 \\
 \cline{2-6}
   & TGSD G+D & 0.01 & 0.1 & 10 & 30 
 \\
 \hline
 \multirow{2}{*}{Bike-interpolate} &  TGSD G+D  & 0.01  & 0.01 & 10 & 25
\\
 \cline{2-6}
  & TGSD G+S   & 0.01  & 0.01 & 10 & 30 
\\
 \hline
  Traffic-impute & TGSD G+R & 0.1 & 0.1 & 10 & 5 
\\
 \hline
Traffic-interpolate & TGSD G+S & 1 & 1 & 10 & 5
\\
 \hline
 Reddit-sp-clustering & TGSD G+D & 1 & 1 & NA & 6
\\
 \hline
   Reddit-sp-k estimation & TGSD G+D & 1 & 1 & 10 & Vary
\\
 \hline
 Reddit-epi-clustering & TGSD G+D & 2 & 0.01 & NA & 25
\\
 \hline
 Reddit-epi-k estimation & TGSD G+D & 2 & 0.01 & 10 & Vary
\\
 \hline
  Reality Mining-clustering & TGSD G+D & 5 & 3 & NA & 5
\\
 \hline
 Reality Mining-k estimation & TGSD G+D & 5 & 3 & 10 & Vary
\\
 \hline

\end{tabular}
 \caption{\footnotesize Parameters for \ourmeth}
\label{table:tgsd_parm_resut}\vsa\vsa\vsa\vsa\vsb
\end{table}

\noindent\textbf{$\bullet$ Clustering (Sec \ref{clustering}).}
For CCTN we use the default parameter $\lambda$ value of $2$ provided by the authors and for the dimensionality of the embeddings we report results for both the default parameter $3$ and when the dimensionality is set to the ground truth number of clusters in each dataset (Tbl.~4 in the main paper). CCTN's limited scalability made grid search not feasible on our large datasets ( it requires close to 5 hours for a single run on the Reddit-sp). For PCA we set the number of dimensions equal to the ground truth number of clusters. 

\noindent\textbf{$\bullet$ Period detection (Sec \ref{Period detection}).}
We set NPM's only parameter---the maximum period in the Ramanujan dictionary---to $50$. AUTO's maximum period is set to $100$. FFT is parameter-free. 

\begin{figure} [t]
    \centering
    \subfigure [$\lambda_1$ v.s. $k$]
    {
        \includegraphics[width=0.23\textwidth]{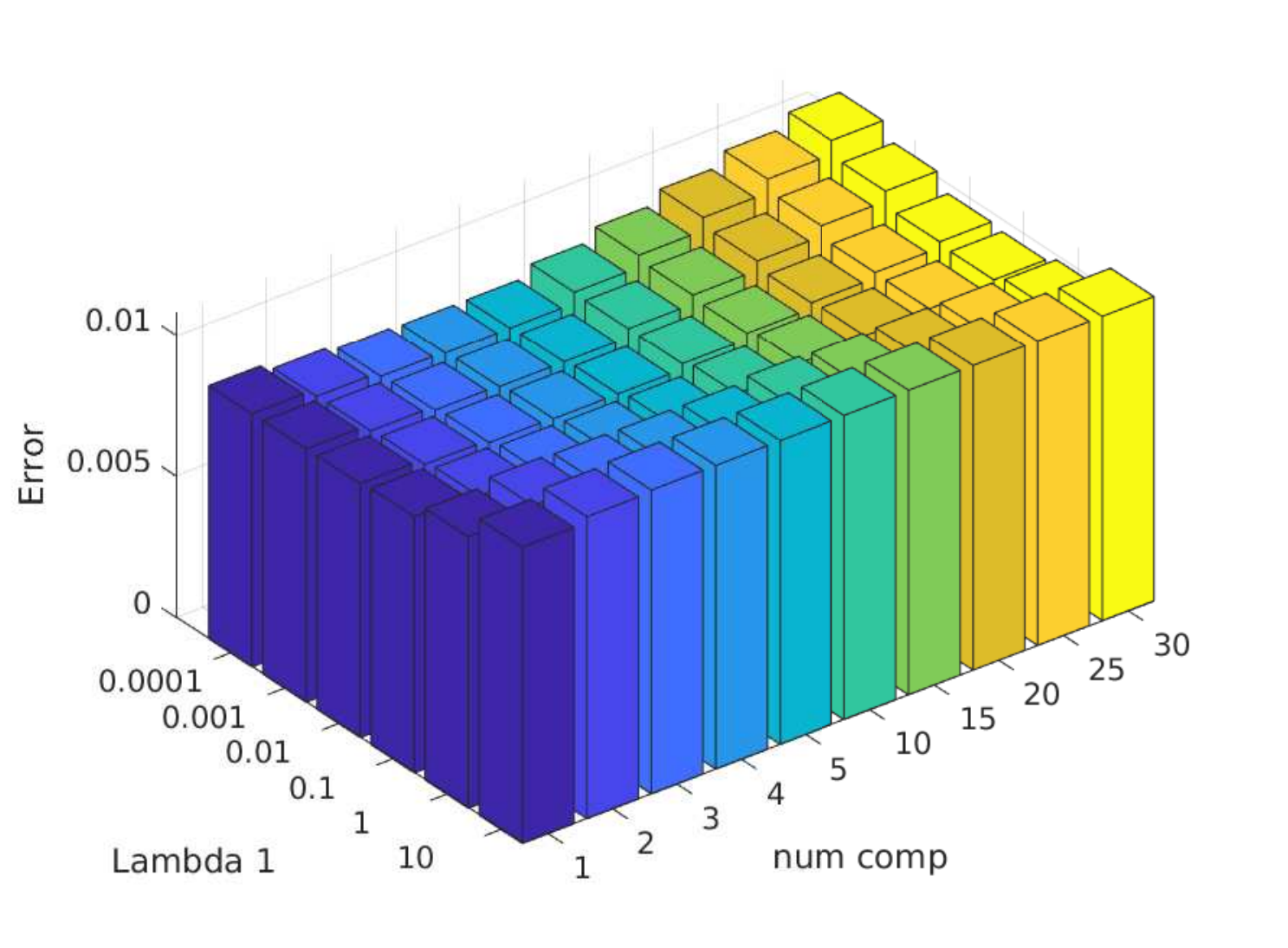}
        \label{fig:param_comp1_2}
    }\hspace{-0.1in}
        \subfigure [$\lambda_1$ v.s. $\lambda_2$]
    {
        \includegraphics[width=0.23\textwidth]{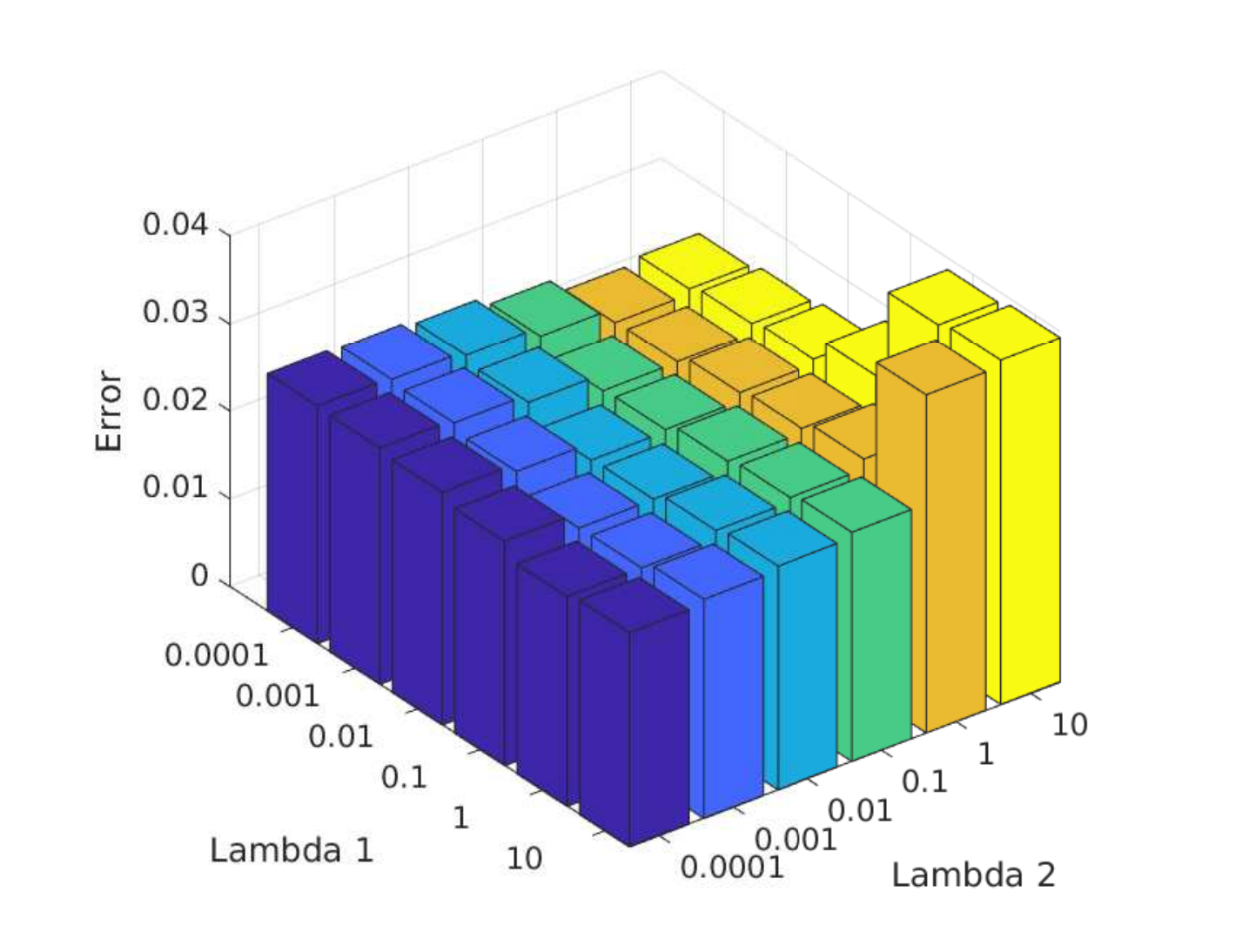}
        \label{fig:param_comp1_k}
    }
           
    

    \caption{Parameter sensitivity analysis for 25\% random missing imputation in the \textit{Synthetic}
    }
    \label{fig:param_sensitive}\vsa\vsa\vsa
\end{figure}



\begin{table*}[th]
\footnotesize
\setlength\tabcolsep{3 pt}
\centering
 \begin{tabular}{|l|c|c|c|c|c|c|c|c|} 
  \hline
    & \multicolumn{3}{|c|}{BRITS} & \multicolumn{3}{|c|}{MCG}    \\
     \hline
  & batch size & epoch  & hidden layer size   & Nuclear norm weight & Column graph weight  & Row graph weight  \\
  \hline
 Synthetic-impute  & 54 & 1000 & 128 & 0.1 & 1 & 0.01 
 \\
 \hline
Synthetic-interpolate  & 108 & 500 & 128  & 0.1 & 0.01 & 0.1 \\
 \hline
 Bike-impute  & 216 & 500 & 64 & 1 & 0.01 & 0.01
 \\
 \hline
Bike-interpolate & 216 & 500 & 128 & 0.5 & 1 & 0.01
\\
 \hline
  Traffic-impute  &  54 & 1000 & 34 & 0.1 & 0.01 & 0.01
\\
 \hline
Traffic-interpolate & 54 & 500 & 128 & 0.1 & 0.01 & 0.01 \\
 \hline
 & \multicolumn{3}{|c|}{GEMS-HD+}  & \multicolumn{3}{|c|}{LRDS}
  \\
   \hline
    & Dictionary Size & Target Atom sparsity  & Target Signal Sparsity   & Nuclear norm weight & Smoothness weight  & Convergence weight  \\
  \hline
 Synthetic-impute  & 5 & 30  & 15 & 0.1 & 0.1 & 0.1 
 \\
 \hline
Synthetic-interpolate  & 5 & 5 & 5 & 0.1 & 1 & 0.1 
\\
 \hline
 Bike-impute  & 20 & 30 & 30 & 1 & 0.1 & 1 
 \\
 \hline
Bike-interpolate  & 5 & 5 & 5  & 1 & 0.1 & 0.1 
\\
 \hline
  Traffic-impute  & 5 & 15 & 30 & 0.1 & 0.5 & 1
\\
 \hline
Traffic-interpolate & 5 & 5 & 5 & 0.1 & 1 & 0.1 \\
 \hline
\end{tabular}
 \caption{\footnotesize \footnotesize  Optimal parameters found by grid search for imputation and interpolation tasks.}
\label{table:grid_resut}\vsa\vsa
\end{table*}

\begin{table*}[th]
\footnotesize
\setlength\tabcolsep{3 pt}
\centering
 \begin{tabular}{|l|c|c|c|c|c|c|c|c|c|c|c|c|c|} 
  \hline
    & \multicolumn{3}{|c|}{BRITS} & \multicolumn{3}{|c|}{MCG} & \multicolumn{3}{|c|}{LRDS} & \multicolumn{3}{|c|}{GEMS-HD+}   \\
     \hline
  & Avg $\pm$  STD  & Min & Max  &  Avg $\pm$  STD   & Min & Max  & Avg $\pm$  STD  & Min & Max  &  Avg $\pm$  STD & Min & Max  \\
  \hline
 Synthetic-impute  & 0.0205 $\pm$ 0.0029  &  0.0193 & 0.0394 & 0.0154 $\pm$  0.0006  & 0.0144 & 0.0164 &   0.0152 $\pm$ 0.0008  &     0.0145 & 0.0162 & 0.0165 $\pm$ 0.0008 & 0.0152 &  0.0182
 \\
 \hline
Synthetic-interpolate  & 0.0167 $\pm$ 0.0054  &  0.0152 & 0.0554  & 0.0173 $\pm$ 0.0015 & 0.0155 &  0.0220  & 0.0212 $\pm$ 0.0013  & 0.0193 &  0.0229  & 0.0167 $\pm$ 0.0000 & 0.0167 & 0.0167\\
 \hline
 Bike-impute  &  0.0346 $\pm$ 0.0017 &   0.0324 & 0.0422 & 0.0307 $\pm$  0.0026  & 0.0265 & 0.0347 &   0.0162 $\pm$  0.0001 & 0.0160 &   0.0164  & 0.0174 $\pm$  0.0013 & 0.0160 & 0.0218
 \\
 \hline
Bike-interpolate  &  0.0352 $\pm$  0.0021 &  0.0329 & 0.0449  & 0.0320 $\pm$  0.0016  & 0.0297 &  0.0347  &  0.0335 $\pm$ 0.0002  & 0.0331 & 0.0337  & 0.0203  $\pm$  0.0000 & 0.0203  & 0.0203
\\
 \hline
  Traffic-impute  &  0.0110 $\pm$  0.00005 &   0.0110 & 0.0112  & 0.0025 $\pm$  0.0000 & 0.0025 & 0.0025  & 0.0112 $\pm$  0.0047 & 0.0053 & 0.0167 &  0.0083 $\pm$ 0.0004 & 0.0076 &  0.0091 
\\
 \hline
Traffic-interpolate & 0.0138 $\pm$  0.0089 &   0.0097 & 0.0340 & 0.0160 $\pm$  0.0000 & 0.0160 & 0.0160 & 0.0702 $\pm$ 0.0004 &   0.0694 & 0.0706  &  0.0111 $\pm$ 0.000 & 0.0111 & 0.0111 \\
 \hline

\end{tabular}
 \caption{\footnotesize \footnotesize Average, standard deviation, minimum, and maximum RMSE found in grid search for each data set and competing method in imputation and interpolation experiments.}
\label{table:grid_stat}\vsa\vsa
\end{table*}


\noindent{\bf $\bullet$ Parameter search and optimal parameters for \ourmeth.}
We performed grid search for all variations of \ourmeth for the graph signal decomposition, imputation, and interpolation experiments (detailed in Tbl.\ref{table:grid_all}).
For period detection we set $K$ =1, $\lambda_1=$ .1, and
$\lambda_2=0.1$. For the estimation of $k$ we set $\lambda_1=0.01$, $\lambda_2=0.001$. The settings for \ourmeth in all experiments except graph signal decomposition can be found in Tbl. \ref{table:tgsd_parm_resut}. 

While we can employ cross-validation to learn good regularization parameters values by occluding values from $X$ (akin to our approach for $k$ estimation), we are also interested in characterizing the sensitivity of \ourmeth to these parameters. In Fig.~\ref{fig:param_sensitive},
we present the sensitivity of our model to pairs of its key hyper-parameters $\left \{ \lambda_1, \lambda_2, k\right \}$. We fix one parameter and vary the other two, and
present the random value imputation error at $25\%$ missing values for the \textit{Synthetic} data set and \ourmeth G+D. Our model's quality deviates from optimal for high $\lambda_1$ and $\lambda_2$. Pushing these values too high forces \ourmeth to overly sparse encodings which ``underfit'' the data. Reasonably small values in the 0.1 range produce optimal performance and the quality is not sensitive to variations in the vicinity of this value. 

\noindent{\bf $\bullet$ Spline imputation.}
For baselines which do not handle missing values directly we utilized spline imputation as a preprocessing step. To this end, we employ Matlab's "interp1" function. 
We apply this imputation method to each node timeseries with missing values to reconstruct the latter. 

\noindent{\bf $\bullet$ An additional baseline of inconclusive performance.}
We also evaluated the Geometric Matrix Completion with Recurrent Multi-Graph Neural Networks (MCGNN)~\cite{MCGNN} as a potential baseline for missing value imputation and temporal interpolation.
This work can be intuitively considered as a deep learning extension to~\cite{kalofolias2014matrix}. 
Unfortunately, we were not able to obtain competitive results from MCGNN (at least an order of magnitude worse SSE than all other competing techniques). Hence we do not report it in the experimental results. We tested MCGNN with the same row and column graphs that we adopt for MCG~\cite{kalofolias2014matrix} on the synthetic $25\%$ random missing task with grid search over the parameters in Tbl.~\ref{table:MCG-grid}. Note, that MCGNN is a deep learning method and the grid search we performed required substantial computational time (over a week).

\begin{table}[h]
\footnotesize
\setlength\tabcolsep{3 pt}
\centering
 \begin{tabular}{|l|c|c|} 
 \hline
 Order of row Chebyshev polynomial & \{4,5,6\} \\
 \hline
 Order of column Chebyshev polynomial &\{4,5,6\} \\
  \hline
  Diffusion steps  &  \{24,32,42\} \\
   \hline
Number of convolution features & \{5,10,15\} \\
\hline
\end{tabular}
\caption{\footnotesize Parameter search for MCGNN\cite{MCGNN}}
\label{table:MCG-grid}\vsa\vsa
\end{table}

While we were not able to obtain competitive results, it is possible that the method can be competitive if larger grid search is employed. A possible limitation could be that the column graph may be ineffective in capturing temporal patterns in the convolution architecture. We refrain from making any stronger conclusions based on the experiments we have performed.

\noindent{\bf $\bullet$ Code.} An implementation of \ourmeth is available at the authors' website: \url{http://www.cs.albany.edu/~petko/lab/code.html}

\end{document}